\documentclass[lettersize,journal]{IEEEtran}

\usepackage{amsmath,amsfonts}
\usepackage{textcomp}

\usepackage{graphicx}
\usepackage{array}
\usepackage{multirow}
\usepackage{booktabs}
\usepackage{tabularx}
\usepackage{stfloats}
\usepackage{capt-of}
\usepackage[caption=false,font=normalsize,labelfont=sf,textfont=sf]{subfig}
\usepackage{caption}

\usepackage{algorithm}
\usepackage{algpseudocode}

\usepackage[table]{xcolor}
\usepackage{colortbl}
\usepackage[most]{tcolorbox}

\usepackage{tikz}
\usepackage{tikzpagenodes}
\usepackage{eso-pic}
\usepackage{fancyhdr}
\usepackage{lastpage}

\usepackage{url}
\usepackage{verbatim}
\usepackage{cite}

\usepackage{pifont}
\usepackage{makecell}
\newcommand{\cmark}{\ding{51}}
\newcommand{\xmark}{\ding{55}}

\usepackage[hidelinks]{hyperref}

\newcolumntype{L}[1]{>{\raggedright\arraybackslash}p{#1}}
\newcolumntype{C}[1]{>{\centering\arraybackslash}p{#1}}

\newif\ifreviewversion
\reviewversionfalse

\fancypagestyle{reviewpage}{
	\fancyhf{}
	\fancyhead[L]{\footnotesize Page \thepage\ of \pageref{LastPage}}

}

\fancypagestyle{plain}{
	\fancyhf{}
	\fancyhead[L]{\footnotesize Page \thepage\ of \pageref{LastPage}}

}

\newcommand{\reviewruler}{%
	\ifreviewversion
	\begin{tikzpicture}[remember picture,overlay]
		\foreach \n [evaluate=\n as \y using {(\n-1)*11.8}] in {1,...,60}{
			\node[
			anchor=east,
			font=\normalfont\footnotesize\sffamily,
			text=black
			] at ([xshift=-9mm,yshift=-\y pt]current page text area.north west) {\n};
		}
	\end{tikzpicture}%
	\fi
}

\AddToShipoutPictureFG{\reviewruler}

\usepackage{zref-savepos}   
\usepackage{xparse}         
\usetikzlibrary{calc}       

\newcounter{revtagcounter}
\newlength{\revtagoutersep}
\newlength{\revtagyshift}
\ExplSyntaxOn
\cs_new_protected:Npn \rev_tag_stack:n #1
{
	\seq_clear:N \l_tmpa_seq
	\clist_map_inline:nn {#1}
	{
		\seq_put_right:Nn \l_tmpa_seq
		{
			\textcolor{blue}{\normalfont\bfseries\itshape\normalsize ##1}
		}
	}
	\begin{tabular}{@{}c@{}}
		\seq_use:Nn \l_tmpa_seq {\\[0.1ex]}
	\end{tabular}
}
\NewDocumentCommand{\revtagstack}{m}{\rev_tag_stack:n {#1}}
\ExplSyntaxOff

\newcommand{\revmargintag}[1]{%
	\ifreviewversion
	\stepcounter{revtagcounter}%
	\zsavepos{revpos-\therevtagcounter}%
	\tikz[remember picture,overlay,baseline]{%
		\coordinate (revcoord-\therevtagcounter) at (0,0);
	}%
	\ifdim\zposx{revpos-\therevtagcounter}sp < .5\paperwidth
	\tikz[remember picture,overlay]{%
		\node[
		anchor=east,
		inner sep=0pt
		] at ([xshift=-\revtagoutersep,yshift=\revtagyshift]
		current page text area.west |- revcoord-\therevtagcounter)
		{\revtagstack{#1}};
	}%
	\else
	\tikz[remember picture,overlay]{%
		\node[
		anchor=west,
		inner sep=0pt
		] at ([xshift=\revtagoutersep,yshift=\revtagyshift]
		current page text area.east |- revcoord-\therevtagcounter)
		{\revtagstack{#1}};
	}%
	\fi
	\fi
}

\NewDocumentCommand{\Rev}{m +m}{%
	\ifreviewversion
	\revmargintag{#1}%
	{\color{blue}#2}%
	\else
	#2%
	\fi
}

\NewDocumentEnvironment{revise}{m +b}
{%
	\Rev{#1}{#2}%
}
{}

\DeclareCaptionFont{reviewblue}{\color{blue}}

\NewDocumentEnvironment{reviewbluetable}{m +b}
{%
	\ifreviewversion
	\begingroup
	\revmargintag{#1}
	\color{blue}
	\arrayrulecolor{blue}
	\captionsetup{
		labelfont={sf,reviewblue},
		textfont={sf,reviewblue}
	}
	#2%
	\arrayrulecolor{black}
	\endgroup
	\else
	#2%
	\fi
}
{}

\usepackage{etoolbox}

\makeatletter

\newcommand{\BibRevTag}[2]{%
	\expandafter\gdef\csname bibrevtag@#1\endcsname{#2}%
}

\newcommand{\PrintBibRevTag}[1]{%
	\ifreviewversion
	\@ifundefined{bibrevtag@#1}%
	{}%
	{\revmargintag{\csname bibrevtag@#1\endcsname}}%
	\fi
}

\apptocmd{\@bibitem}
{\PrintBibRevTag{#1}}
{}
{\typeout{Failed to patch \string\@bibitem}}

\apptocmd{\@lbibitem}
{\PrintBibRevTag{#2}}
{}
{\typeout{Failed to patch \string\@lbibitem}}

\makeatother

\begin{document}

\title{DifferAD-R1: A Difference-Guided Industrial Anomaly Localization with Multimodal Large Language Models }

\author{
	Dingrong~Wang,
	Xian~Tao,
	Zhen~Qu,
	Hengliang~Luo,
	Xinyi~Gong,
	Fei~Shen,
	Zhengtao~Zhang and Guiguang~Ding%
	\thanks{This work was supported in part by the project ZR2025LGY002 supported by Shandong Provincial Natural Science Foundation, in part by the National Natural Science Foundation of China under Grants 62373350 and       62371179, in part by the Youth Innovation Promotion Association of the Chinese Academy of Sciences (CAS) under Grant 2023145, and in part by the Beijing Nova Program under Grant 20240484687.(Corresponding author: Xian Tao)
		
		Dingrong Wang, Zhen Qu, Fei Shen, and Zhengtao Zhang are with the Institute of Automation, Chinese Academy of Sciences (CAS), Beijing 100190, China, and also with the School of Artificial Intelligence, University of Chinese Academy of Sciences, Beijing 100049, China. 
		
		Xian Tao is with the Institute of Automation, CAS, Beijing 100190, China, CASI Vision Technology Co., Ltd., Luoyang 471000, China and also with the Shandong Laboratory of Aluminum Advanced Manufacturing in Binzhou (SLAAMB), Binzhou Institute of Technology, Weiqiao-UCAS Science and Technology Park, Binzhou 256606, China (Corresponding Author Email: taoxian2013@ia.ac.cn). 
		
		Hengliang Luo is with CASI Vision Technology Co., Ltd., Luoyang 471000, China.
		
		Xinyi Gong is with the Space Information Research Institute, Hangzhou Dianzi University, Hangzhou 310018, China. 
		
		Guiguang Ding is with the School of Software, Tsinghua University, Beijing 100084, China.}%
}

\markboth{Journal of \LaTeX\ Class Files,~Vol.~14, No.~8, August~2021}%
{Shell \MakeLowercase{\textit{et al.}}: A Sample Article Using IEEEtran.cls for IEEE Journals}


\maketitle

\thispagestyle{reviewpage}
\pagestyle{reviewpage}

\begin{abstract}
Industrial anomaly localization aims to accurately identify and localize abnormal regions in industrial products, addressing the critical challenge of detecting unseen defect categories in real-world scenarios. Traditional closed-set methods often suffer from poor cross-scenario generalization, while existing Multimodal Large Language Model (MLLM)-based approaches face two core limitations: they either adopt QA-style paradigms misaligned with the practical demands of localization, or rely on standard optimization techniques such as Group Relative Policy Optimization (GRPO), which fails to deliver effective learning signals for subtle defects. To tackle these issues, this paper proposes DifferAD-R1, an MLLM-augmented reinforcement learning framework tailored for industrial anomaly localization. We design a Difference-Guided dual-image paradigm, which reformulates the localization task as a one-shot difference grounding problem to effectively explore cross-scenario anomalies. A Dual-Consistency Localization Reward is developed for hard-to-detect anomalies, enhancing optimization stability and robustness. Additionally, we integrate a difficulty-aware strategy with adaptive reweighting and group-wise resampling to prioritize learning on challenging instances. To facilitate evaluations in real-world industrial settings, we construct the AD-DualDiff dataset, comprising 13K paired images across 20 categories. Experimental results demonstrate that DifferAD-R1 significantly outperforms existing baselines and achieves competitive performance compared to large-scale models like Qwen3-VL (235B parameters). Our code is publicly available at: https://github.com/Rong2026/work-1. 
\end{abstract}


\section{Introduction}
\begin{figure*}[!t]
	\centering
	\includegraphics[width=\textwidth]{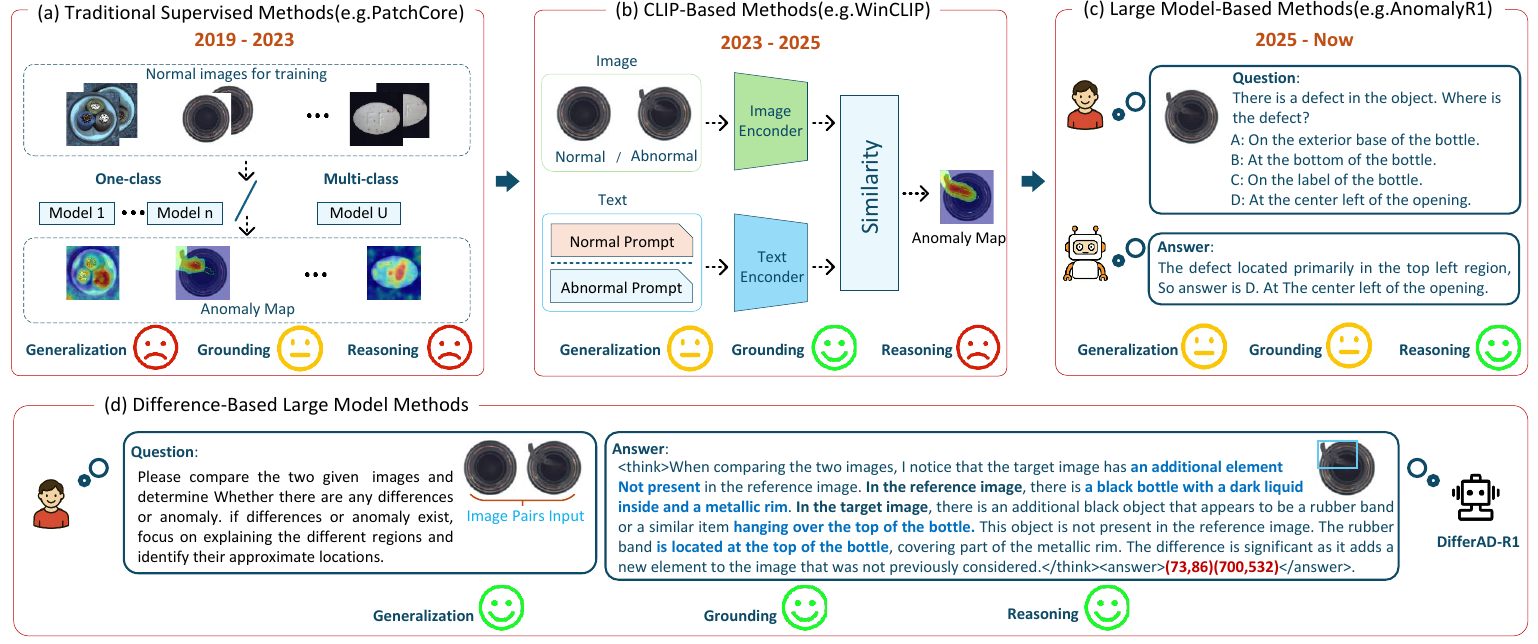} 
	\caption{An overview of industrial anomaly detection paradigms, (a) traditional supervised methods to (b) CLIP-based approaches and (c) large-model QA formulations, contrasted with (d) our difference-guided large vision--language framework.}
	\label{fig:history}
\end{figure*}

\IEEEPARstart{I}{ndustrial} anomaly detection (IAD) is a foundational component of modern manufacturing, critical to safeguarding product quality and optimizing manufacturing yield. As illustrated in Fig.~\ref{fig:history}(a), early research in this field has predominantly been framed under one-class or multi-class learning paradigms~\cite{defard2021padim,zavrtanik2021draem,pirnay2022inpainting,roth2022patchcore}, where models are trained to distinguish normal from abnormal samples within a predefined set of product categories. These methods typically leverage three core technical routes: image/feature reconstruction (e.g., DRAEM~\cite{zavrtanik2021draem}, Dual-Siamese~\cite{tao2022unsupervised}, RealNet~\cite{zhang2024realnet}), feature distribution modeling (e.g., PaDiM~\cite{defard2021padim}, PatchCore~\cite{roth2022patchcore}, CFA~\cite{lee2022cfa}), and knowledge distillation (e.g., REASON~\cite{wang2025Distillation1}, Pull \& Push~\cite{zhou2022Distillation2}, UniSTAD~\cite{liu2024unistad3}). However, these formulations implicitly assume that training and testing data follow identical data distributions, a premise that rarely holds in real industrial environments, where unseen defect types inevitably arise due to dynamic production conditions (e.g., material variations or process adjustments). This closed-world assumption thus leads to significant performance degradation in real-world deployments.

To mitigate the limitations of closed-set paradigms, WinCLIP~\cite{jeong2023winclip} pioneered the integration of CLIP-based contrastive learning into zero-shot/few-shot anomaly detection in 2023. As illustrated in Fig.~\ref{fig:history}(b), these methods leverage image--text alignment to identify anomalous patterns without the need for category-specific training. Subsequent extensions, including AnomalyCLIP~\cite{zhou2023anomalyclip}, VCP-CLIP~\cite{qu2024vcp}, and FiLo~\cite{gu2024filo}, have further advanced semantic understanding of anomalies by designing sophisticated textual prompts and increasingly complex alignment mechanisms. Meanwhile, pixel-level contrastive learning has also shown its effectiveness in dense visual prediction~\cite{wu2023transformer}. Nevertheless, these contrastive learning-based approaches still face two key limitations for practical industrial localization: first, they fail to fully exploit the rich semantic information inherent to industrial anomalies; second, they lack the ability to integrate inherent reasoning capabilities required for precise localization, as their design is overly reliant on contrastive learning paradigms without leveraging advanced cognitive mechanisms.

Recently, Multimodal Large Language Models (MLLMs) have witnessed unprecedented advancements, with models such as LLaVA-OneVision-1.5~\cite{an2025llavaonvevision}, Qwen3-VL~\cite{Bai2025Qwen3VLTR}, and InternVL3.5~\cite{wang2025internvl35} demonstrating state-of-the-art performance across tasks including summarization, visual localization, and visual question answering (VQA). Driven by these breakthroughs, an increasing number of works have sought to adapt MLLMs for industrial anomaly detection~\cite{zhang2024gpt,chen2025can,kwon2025logicqa,xu2025plovad,li2026semaero}. As illustrated in Fig.~\ref{fig:history}(c), most existing methods directly input images into MLLMs to generate textual descriptions or select predefined answer choices. However, such QA-style paradigms lack fine-grained visual perception capabilities and instead rely on aligning textual prompts with input images to infer anomalies. Consequently, this form of low-level visual understanding struggles to meet the demands of anomaly detection in complex real-world industrial scenarios.
To address this issue, recent works, such as EMIT~\cite{guan2025emit}, AnomalyR1~\cite{chao2025anomalyr1}, and OmniAD~\cite{zhao2025omniad}, have adopted Group Relative Policy Optimization (GRPO)~\cite{shao2024grpo} to align MLLMs with downstream tasks through visual-guided textual reasoning. Nevertheless, when facing challenging cases—such as subtle defects embedded in complex industrial backgrounds—standard GRPO~\cite{shao2024grpo} often fails to deliver effective learning signals, leading to inaccurate and unstable anomaly localization.

Furthermore, the capability to detect anomalies across diverse defect types and cross-scenario settings is far more pivotal in real-world industrial quality inspections. Most conventional anomaly detection methods have already reached near-saturated performance on fixed benchmark datasets such as MVTec AD~\cite{bergmann2019mvtec}, with image-level and pixel-level AUROC scores frequently exceeding 99\%~\cite{roth2022patchcore,guo2025Dinomaly,liu2024unistad3}. While recent works have sought to integrate Multimodal Large Language Models (MLLMs) into anomaly detection by leveraging open datasets, such as MMAD~\cite{jiang2024mmad}, InstructIAD~\cite{li2025triad}, and Anomaly-Instruct-125k~\cite{xu2025anomalyov}, these approaches predominantly frame the task as a question-answering (QA) paradigm, where the model selects predefined answers instead of explicitly localizing defective regions. Such QA-style paradigms are inherently misaligned with the practical demands of industrial quality inspection, where precise spatial localization of anomalies is a core requirement for downstream quality control and process optimization. Consequently, there is an urgent demand for novel modeling paradigms and datasets that more faithfully mirror real-world industrial anomaly localization scenarios.

To tackle the aforementioned challenges, we propose DifferAD-R1, an MLLM-augmented reinforcement learning framework for industrial anomaly localization (Fig.~\ref{fig:history}(d)). Specifically, we design a Difference-Guided dual-image paradigm that reformulates anomaly localization as a one-shot difference grounding task, enabling high-level visual reasoning and direct prediction of anomaly bounding boxes. To address the limitations of standard GRPO~\cite{shao2024grpo} in hard anomaly localization scenarios (e.g., subtle defects in complex industrial backgrounds), we develop a Dual-Consistency Localization Reward to provide meaningful learning signals—guiding predicted bounding boxes toward the groundtruth even in the absence of initial overlap. Furthermore, we integrate a difficulty-aware training strategy that prioritizes hard samples via adaptive advantage reweighting, and applies group-wise resampling when the response-level consensus results in zero-advantage signals. To facilitate performance evaluation in real-world industrial settings, we construct the AD-DualDiff dataset, comprising 13K paired images across 20 product categories. Experimental results demonstrate that DifferAD-R1 significantly outperforms existing baselines and achieves competitive localization performance even compared to large-scale models (e.g., Qwen3-VL with 235B parameters~\cite{Bai2025Qwen3VLTR}), highlighting the effectiveness of our proposed framework for industrial anomaly localization.

Building upon the aforementioned motivations, we summarize our key contributions as follows:

\begin{itemize}
	\item \textbf{(Framework)}
	We propose DifferAD-R1, the first MLLM-augmented difference-guided dual-image vision-language framework that explicitly reformulates industrial anomaly localization as a one-shot paired-image difference grounding task, enabling inherent cross-scenario generalization.
	
	\item \textbf{(Strategy)}
	We design a task-adaptive reinforcement learning optimization strategy, integrating a Dual-Consistency Localization Reward and a Difficulty-Aware Sampling Mechanism. This strategy delivers stable and discriminative learning signals even for highly inaccurate initial localization predictions.
	
	\item \textbf{(Dataset)}
	We curate AD-DualDiff, a high-quality real-world industrial anomaly localization dataset, comprising 13K precisely aligned reference--target image pairs spanning 20 industrial product categories, faithfully reflecting practical inspection scenarios.
	
	\item \textbf{(Performance)}
	Our method achieves state-of-the-art (SOTA) localization performance on both AD-DualDiff and the classic MVTec AD benchmark, while demonstrating competitive results against large-scale MLLMs (e.g., Qwen3-VL with 235B parameters). This validates the efficacy of difference-guided vision-language reasoning and reinforcement learning in addressing core challenges of real-world industrial anomaly localization.
\end{itemize}

\section{Related Work}
\begin{figure*}[!t]
	\centering
	\includegraphics[width=\textwidth]{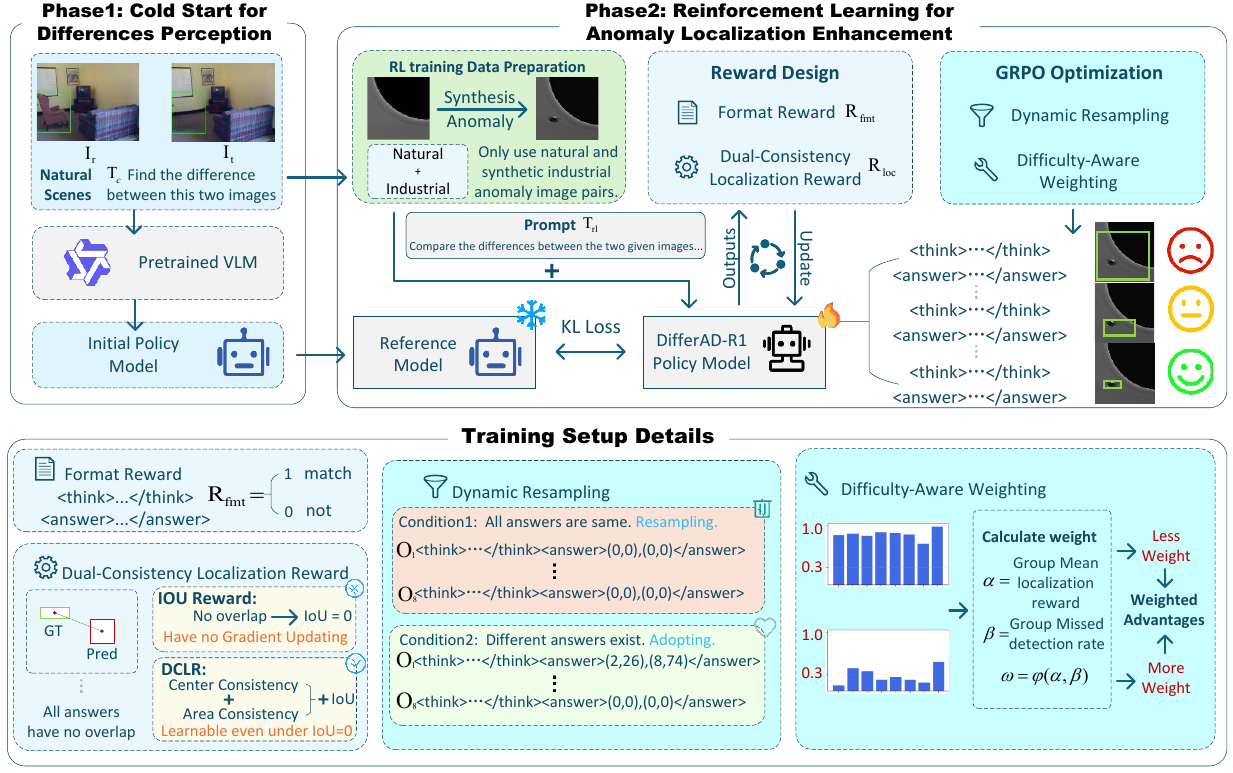} 
	\caption{\Rev{2-5}{Overall pipeline of DifferAD-R1. (a) The framework first builds cross-image difference perception through cold-start training, (b) and then enhances industrial anomaly localization via GRPO with (c) the proposed Dual-Consistency Localization Reward, group-wise resampling, and difficulty-aware weighting.}}
	\label{fig:pipeline}
\end{figure*}

\subsection{Traditional Industrial Anomaly Detection}
Traditional industrial anomaly detection has been predominantly framed under one-class and multi-class learning paradigms, where models distinguish normal from abnormal states within a predefined product category.Early reconstruction-based approaches~\cite{tao2022zhongshu}, including autoencoders, variational autoencoders, and generative adversarial network (GAN) variants, seek to model the distribution of normal patterns and identify anomalies by quantifying reconstruction errors. However, these methods often suffer from the ``identical shortcut'' problem—where both anomaly-free and anomalous samples can be effectively reconstructed during inference, leading to ambiguous error signals and compromised detection performance. To mitigate these limitations, embedding-based methods have emerged as the dominant paradigm. Approaches such as PaDiM~\cite{defard2021padim}, PatchCore~\cite{roth2022patchcore}, and teacher--student distillation frameworks~\cite{bergmann2020STeacher,wang2025Distillation1,guo2025Dinomaly} leverage pretrained feature extractors and characterize normality by modeling the distribution of deep visual embeddings. These methods have delivered state-of-the-art performance on standard benchmarks, including MVTec AD~\cite{bergmann2019mvtec} and VisA~\cite{zou2022visa}. Despite their strong empirical results, embedding-based methods fundamentally frame anomalies as distributional outliers in the feature space, lacking explicit mechanisms to support open-world detection (for unseen defect types) and relational reasoning. WinCLIP~\cite{jeong2023winclip} pioneered the integration of CLIP~\cite{clip} into zero-shot/few-shot anomaly detection by measuring the similarity between manually crafted textual prompts and image features. \Rev{0-1}{Recent zero-/few-shot anomaly detection methods, such as FiLo++~\cite{gu2026filo++}, improve industrial anomaly detection by combining fused fine-grained textual descriptions with deformable localization for more accurate anomaly recognition and localization.} To mitigate reliance on manually designed prompts, AnomalyCLIP~\cite{zhou2023anomalyclip} learns object-agnostic prompt templates to enhance the robustness of zero-shot anomaly detection. VCP-CLIP~\cite{qu2024vcp} further improves textual representations via the integration of visual context, boosting performance in zero-shot anomaly segmentation. More recently, Bayes-PFL~\cite{qu2025bayesian} models the prompt space as a learnable probabilistic distribution, leveraging cross-modal interactions to better align fine-grained visual features with dynamic textual prompts. While CLIP-based methods effectively support zero-shot or few-shot settings, they primarily frame anomaly detection as an image--text similarity scoring task. This formulation renders precise anomaly localization heavily reliant on exquisitely designed prompts and complex cross-modal alignment mechanisms—limitations that hinder their practical deployment in industrial inspection scenarios.

\subsection{Multimodal Large Models for Anomaly Detection}
Early studies on anomaly detection directly adopt pretrained Vision--Language Models (VLMs) without fine-tuning~\cite{chen2025can,kwon2025logicqa,zhang2024gpt}, relying solely on prompting strategies or manually designed visual cues. However, due to the significant domain gap between industrial imagery and generic pretraining data, these zero-fine-tuning approaches fail to capture nuanced defects, limiting their practical applicability in industrial scenarios. To bridge this gap and enhance task alignment, several works leverage supervised fine-tuning (SFT) for VLMs. AnomalyGPT~\cite{gu2024anomalygpt} constructs synthetic defect datasets and combines image decoders with prompt learners to adapt VLMs to anomaly detection tasks. Myriad~\cite{li2023myriad} integrates pretrained industrial anomaly detection (IAD) models as visual experts, providing pseudo-priors to guide the multimodal model’s response generation. Anomaly-OV~\cite{xu2025anomalyov} proposes a ``Look Twice'' feature matching mechanism to adaptively highlight abnormal visual tokens. Despite the improved task specialization brought by SFT, this paradigm relies heavily on high-quality training data and complex visual-expert architectures, hindering end-to-end optimization and generalization to unstructured real-world defects.
Recent efforts have explored reinforcement learning (RL)-based alignment, particularly Group Relative Policy Optimization (GRPO)~\cite{shao2024grpo}, to refine VLM behaviors via task-specific rewards. AnomalyR1~\cite{chao2025anomalyr1} introduces an end-to-end MLLM pipeline with the ROMA reward, jointly optimizing reasoning processes and final predictions. LR-IAD~\cite{zeng2025LRIAD} proposes a focal reward that dynamically weights rare answers to mitigate class imbalance in anomaly detection. EMIT~\cite{guan2025emit} boosts few-shot performance through dual-image contrastive embeddings and soft prompting, while OmniAD~\cite{zhao2025omniad} adopts a hybrid SFT+GRPO training strategy. Despite these advances, most RL-based methods still frame the task as a multiple-choice QA task, misaligned with real industrial demands that require precise grounded localization and interpretable results. This fundamental paradigm mismatch prevents their widespread deployment in practical industrial inspection workflows.

\section{Method}

\subsection{Task Definition}
Industrial anomaly localization with MLLMs aims to identify whether an inspected sample contains anomalous regions, and output both their spatial locations and semantic descriptions. Compared with traditional anomaly classification or pixel-level segmentation tasks, practical industrial scenarios prioritize open-world robust anomaly localization and interpretability under distribution shifts. In this work, we focus on the one-shot industrial anomaly localization scenario. Each test case comprises a reference image $I_r$ and a target image $I_t$, where $I_r$ denotes a normal instance (anomaly-free) and $I_t$ may contain anomalous regions. Given the paired images $(I_r, I_t)$, the model is tasked with two core objectives: first, determining the presence of anomalies in $I_t$; second, generating the corresponding bounding box $B$ (for spatial localization) and natural language description $D$ (for semantic interpretation). To fully exploit the cross-image reasoning capabilities of large vision-language models (LVLMs), we reformulate the task as a paired-image, difference-driven grounding problem—aligning with the Difference-Guided paradigm proposed later. For the sake of interpretable evaluation and clear performance benchmarking, we restrict the task to a single-anomaly setting, where each $I_t$ contains at most one anomalous region.

During training, the model is optimized on a training set $\mathcal{D}_{\text{train}} = \{(I_r, I_t)\}$ consisting exclusively of natural scene image pairs and synthetically generated industrial anomaly image pairs—with no real industrial defect samples involved. This design addresses the scarcity of real industrial defect data, a common challenge in practical inspection scenarios. At the test stage, the model is evaluated on a disjoint test set $\mathcal{D}_{\text{test}}$, which includes our proposed AD-DualDiff dataset and MVTec AD. Critically, the training and evaluation datasets satisfy $\mathcal{D}_{\text{train}} \cap \mathcal{D}_{\text{test}} = \emptyset$, thereby enabling a rigorous assessment of the model’s cross-dataset generalization capability—aligning with the core demand for open-world industrial anomaly localization.

\subsection{DifferAD-R1}
To address the limitations of existing industrial anomaly detection methods in localization accuracy, cross-scenario generalization, and interpretability, we propose the DifferAD-R1 framework, as illustrated in Fig.~\ref{fig:pipeline}. \Rev{3-1}{ Unlike single-image anomaly detection or QA-style MLLM methods, DifferAD-R1 reformulates industrial anomaly localization as a reference-guided difference grounding task. Given a normal reference image and a target image, the model directly compares the two images, reasons about meaningful local discrepancies, and grounds the anomalous region.} It adopts a two-stage iterative design: first, a cold-start difference perception stage that builds difference comparison and interpretable reasoning capabilities using only natural image pairs. Second, a subsequent reinforcement learning (RL) stage, where we propose our novel Dual-Consistency Localization Reward and a GRPO-based training strategy to enhance localization refinement—training exclusively on natural image pairs and synthetic industrial anomaly image pairs. This two-stage design enables DifferAD-R1 to localize anomalous regions through both fine-grained perception and cross-image contextual comparison, closely mimicking the way human inspectors identify defects by contrasting test samples with normal references. Such a paradigm effectively bridges the gap between general cross-image reasoning and task-specific industrial anomaly localization.

\subsubsection{\textbf{Cold Start for Difference Perception}}
As illustrated in Fig.~\ref{fig:pipeline}, the cold-start stage is designed to endow the model with generic cross-image difference perception and interpretable reasoning abilities, serving as a foundational pre-adaptation before introducing industrial anomaly-specific supervision. Rather than directly training on industrial anomaly data, we first conduct fine-tuning on large-scale natural image pairs, where the task is framed as identifying and describing cross-image visual differences.

Formally, given a task-specific textual prompt $T_{c}$, a reference image $I_r$, and a target image $I_t$,
the base vision-language model $\mathcal{M}$ is trained to generate a structured output
\begin{equation}
\label{equa1}
(D, B) = \mathcal{M}(T, I_r, I_t).
\end{equation}
where $B = (x_1, y_1),(x_2, y_2)$ denotes the localized region in $I_t$ that differs from $I_r$,
and $D$ is a natural-language description explaining the corresponding visual differences. This structured output aligns with the dual requirements of spatial localization and semantic interpretability defined in the task setup. 

In practice, we adopt UniVG-R1~\cite{bai2025univg} as our cold-start model—a vision-language model pre-aligned for multi-image difference reasoning on natural scenes. Correspondingly, the base model  $\mathcal{M}(\cdot)$ is instantiated with the Qwen2-VL-7B backbone employed in UniVG-R1. Notably, no industrial anomaly data are utilized in this stage; the cold-start model merely provides an initial prior for cross-image comparison, spatial grounding, and difference-aware explanation. This prior, learned from natural image pairs, enables effective transfer to industrial inspection scenarios by framing anomalies as deviations from normal reference states (i.e., $I_r$). Building on this cold-start foundation, the second-stage reinforcement learning (RL) serves as our core task-specific alignment module, which explicitly optimizes anomaly localization and semantic explanation.

\subsubsection{\textbf{Reinforcement Learning for Anomaly Localization Enhancement}}

While the cold-start stage provides structured difference perception and explanation capability, the model remains insufficient to meet practical requirements of real industrial anomaly localization. To enable accurate industrial anomaly localization, we design a reinforcement learning stage to align the model with task-specific localization objectives.

As shown in the Fig.~\ref{fig:pipeline}, the optimization follows GRPO framework.
Since the model generates outputs autoregressively, industrial anomaly localization
can be formulated as a sequential decision-making problem.
Given a textual prompt $T_{o}$ and an image pair $(I_r, I_t)$, the policy $\pi_\theta$
produces a token sequence

\begin{equation}
	\label{equa2}
	\tau = (a_1, a_2, \dots, a_T),
\end{equation}
which is parsed into an anomaly bounding box $B$ and a semantic description $D$.
Reinforcement learning optimizes the expected reward over such sequences,
guiding the model toward accurate and interpretable anomaly localization.
Our goal is to identify the most effective reward and GRPO optimization strategy for DifferAD-R1.

\paragraph{\textbf{Reward Design}}
To effectively align the generated outputs with the localization objective, we design two complementary reward functions: a \emph{format reward} and a \emph{localization reward}.

\paragraph*{Format Reward($R_{fmt}$)}
The format reward enforces structural validity of the generated outputs. Specifically, the model is required to generate results following a predefined template:
\texttt{<think>...</think><answer>...</answer>}.
The output is verified to strictly conform to this template using fault-tolerant regular expression matching. If the generated output strictly conforms to this format, a reward of 1 is assigned; otherwise, the reward is set to 0. 
This simple binary constraint effectively restricts the generation space and stabilizes reinforcement learning (RL), especially in the early training stage, laying a foundation for the subsequent localization reward. 
\paragraph*{Dual-Consistency Localization Reward ($R_{\mathrm{loc}}$)}
To address the difficulty of accurately localizing subtle defects in complex backgrounds, Dual-Consistency Localization Reward is designed to extend the standard IoU-based reward formulation. Given a predicted bounding box $B_p$ and ground-truth box $B_{gt}$, the intersection-over-union (IoU) score is computed as: 

\begin{equation}
	\label{eq:iou}
	\mathrm{IoU}(B_p,B_{gt})
	= \frac{|B_p \cap B_{gt}|}{|B_p \cup B_{gt}|}.
\end{equation}

\begin{figure}[!t]
	\centering
	\includegraphics[width=3.5in]{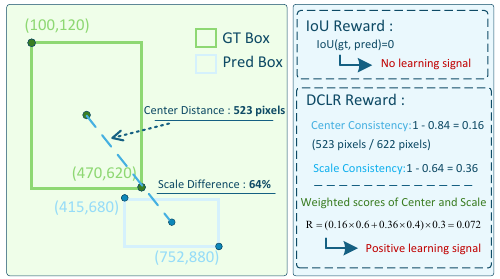} 
	\caption{\Rev{2-5}{
			Illustration of the Dual-Consistency Localization Reward. When the predicted box has no overlap with the ground truth, DCLR still provides positive learning signals through center and scale consistency.
	}}
	\label{fig:DCLR}
\end{figure}

Unlike natural scene large-object recognition, industrial scenarios contain numerous subtle, small-scale anomalies that are inherently challenging to localize. Existing R1-based reasoning grounding models~\cite{bai2025univg} typically adopt IoU-based reward functions. However, in early training stages, the overlap between $B_p$  and $B_{gt}$ is usually low, leading to a lack of effective learning signals (see Fig.~\ref{fig:DCLR}). To resolve this, an enhanced IoU-based reward with geometric consistency optimization is proposed. Its core idea is to construct a weighted IoU reward by jointly modeling center and scale consistency. This mechanism guarantees informative reward signals even with low overlap between $B_p$ and $B_{gt}$.

Specifically, the center consistency reward is characterized by the Euclidean distance between the centers of $B_p$ and ground-truth $B_{gt}$, as illustrated in Fig.~\ref{fig:DCLR}. Its formulation is:
\begin{equation}
	\label{eq:center_cons}
	s_{\mathrm{center}}
	= 1 - \min\!\left(
	1,\,
	\frac{\|c(B_p)-c(B_{gt})\|_2}
	{\max(\mathrm{diag}(B_p),\,\mathrm{diag}(B_{gt}),\,1) + \epsilon}
	\right).
\end{equation}
where $c(B)=\big(\frac{x_1+x_2}{2},\,\frac{y_1+y_2}{2}\big)$ denotes the center coordinate of bounding box
$B=(x_1,y_1,x_2,y_2)$, $\mathrm{diag}(B)$ denotes the Euclidean length of the bounding box diagonal,
and $\epsilon$ is a small constant for numerical stability.

The scale consistency reward is defined based on the relative difference in region area
between the predicted bounding box $B_p$ and the ground-truth bounding box $B_{gt}$,
with the formulation:
\begin{equation}
	\label{eq:scale_cons}
	s_{\mathrm{scale}}
	= 1 - \min\!\left(
	1,\,
	\frac{|A(B_p)-A(B_{gt})|}
	{A(B_{gt}) + \epsilon}
	\right).
\end{equation}
where $A(B)=\max(0,x_2-x_1)\max(0,y_2-y_1)$ denotes the area of a bounding box
$B=(x_1,y_1,x_2,y_2)$, and $\epsilon$ is a small constant for numerical stability.

The dual-consistency score is obtained by combining the center and scale terms,
as follows:
\begin{equation}
	\label{eq:r_cons}
	\begin{aligned}
		S_{\mathrm{cons}}(B_p,B_{gt})
		&= \alpha \, s_{\mathrm{center}}
		+ (1-\alpha)\, s_{\mathrm{scale}}, \\
		&\quad \alpha \in [0,1].
	\end{aligned}
\end{equation}
where $\alpha$ balances the trade-off between center alignment and scale consistency. 

Finally, the overall localization reward $R_{\mathrm{loc}}$ adaptively combines
the IoU term in Eq.~\eqref{eq:iou} and the dual-consistency score in
Eq.~\eqref{eq:r_cons}:
\begin{equation}
	\label{eq:rloc}
	R_{\mathrm{loc}}=
	\begin{cases}
		\mathrm{IoU} + \delta, & \mathrm{IoU} \ge \tau, \\
		\mathrm{IoU} + (1-\mathrm{IoU})\,\lambda\, S_{\mathrm{cons}}, & \mathrm{IoU} < \tau.
	\end{cases}
\end{equation}
where
$\tau$ is the IoU threshold,
$\lambda$ regulates the contribution of the consistency term under low IoU,
and $\delta$ is a small bounded adjustment term. Therefore, the final reward for a response $o_i$ is defined as $R(o_i) = R_{fmt}(o_i) + R_{loc}(o_i)$.

\paragraph{\textbf{GRPO Optimization}}
As shown in the Fig.~\ref{fig:pipeline}, we adopt GRPO to optimize the policy and compute group-wise advantages.
For each input prompt $T_{rl}$, multiple responses are sampled to form a response group.
The advantage of response $o_i$ is computed as
\begin{equation}
	\label{eq:grpo_adv}
	A_i =
	\frac{R(o_i) - \mathrm{mean}\{R(o_j)\}_{j=1}^{G}}
	{\mathrm{std}\{R(o_j)\}_{j=1}^{G}}.
\end{equation}
where $\mathrm{mean}\{R(o_j)\}_{j=1}^{G}$ and $\mathrm{std}\{R(o_j)\}_{j=1}^{G}$ denote the mean and standard deviation of rewards in the group, respectively.

Beyond standard GRPO, two optimization strategies, including Group-wise Resampling and Difficulty-Aware Weighting, are proposed to enhance stability and emphasize challenging anomaly localization cases.

\paragraph*{Group-wise Resampling}
In GRPO training, multiple responses are sampled per input to form a group. When all responses in a group yield nearly identical rewards, the group suffers from degenerate reward variance and provides no effective learning signal. To mitigate this, responses are resampled for the same input until at least one non-trivial localization outcome is observed, as shown in Fig.~\ref{fig:pipeline}(middle of the training setup details ). This strategy avoids degenerate advantage estimates and boosts training stability without extra supervision.

\paragraph*{Difficulty-Aware Weighting}
While group-wise resampling mitigates degenerate reward variance, the optimization process may still be dominated by easy samples, leading to inadequate learning on challenging cases. To address this, as shown in  Fig.~\ref{fig:pipeline} (right of the  training setup details), a difficulty-aware weighting strategy is proposed to explicitly integrate two complementary indicators of sample difficulty.

To better identify hard anomaly samples, two complementary indicators are jointly considered: $\bar{R}_{\mathrm{loc}}^{(g)}$ (Group Localization Quality) measures the overall localization performance of a response group, and $p_{\mathrm{miss}}^{(g)}$ (Group Missed Detection Rate) captures uncertainty and failure frequency within the group. Their formulations are: 

\begin{equation}
	\label{eq:mean_loc_reward}
	\bar{R}_{\mathrm{loc}}^{(g)} = \frac{1}{G} \sum_{j=1}^{G} R_{\mathrm{loc}}^{(j)},
\end{equation}
\begin{equation}
	\label{eq:miss_rate}
	p_{\mathrm{miss}}^{(g)} = \frac{1}{G} \sum_{j=1}^{G}
	\mathbb{I}\!\left(R_{\mathrm{loc}}^{(j)} < \tau \right),
\end{equation}
where \(R_{\mathrm{loc}}^{(j)}\) denotes the localization score of the \(j\)-th response,
\(G\) is the number of sampled responses in group \(g\),
\(\tau\) is a predefined localization threshold,
and \(\mathbb{I}(\cdot)\) denotes the indicator function.

Based on the above two indicators, a difficulty-aware weight $w_{\mathrm{diff}}^{(g)}$ is constructed by jointly
considering localization quality and failure uncertainty. This approach assigns larger weights to harder samples in a smooth and bounded manner,
avoiding abrupt amplification that may destabilize training.
\begin{equation}
	\label{eq:diff_weight}
	w_{\mathrm{diff}}^{(g)} =
	\frac{1}{2}
	\left(
	\exp\!\left(1 - \bar{R}_{\mathrm{loc}}^{(g)}\right)
	+
	\sqrt{1 + p_{\mathrm{miss}}^{(g)}}
	\right).
\end{equation}

Hence, unlike selective reweighting schemes that rely on explicit anomaly indicators,
the difficulty-aware weight $w_{\mathrm{diff}}^{(g)}$ is incorporated into the GRPO
optimization strategy.
This weighting mechanism increases the contribution of challenging samples within each
response group, thereby enhancing the model’s ability to localize subtle anomalies.
Finally, the weighted loss for the $i$-th response is defined as
\begin{equation}
	\label{eq:final_weighted_loss}
	\mathcal{L}_i =
	w_{\mathrm{diff}}^{(g)} \,
	\mathcal{L}_i^{\mathrm{GRPO}},
\end{equation}
where $\mathcal{L}_i^{\mathrm{GRPO}}$ denotes the GRPO per-token loss of the $i$-th response.

\textbf{Discussion.}
The proposed reinforcement learning framework integrates complementary supervision signals
at both the localization and optimization levels.
The \emph{Dual-Consistency Localization Reward} provides dense and continuous geometric guidance,
ensuring informative learning signals even under low- or zero-IoU conditions,
while the \emph{difficulty-aware weighting strategy} adaptively emphasizes challenging response groups
during GRPO optimization.
Together, these designs promote balanced and stable learning dynamics that prioritize
fine-grained anomaly localization without over-amplifying easy samples.

\begin{figure}[t]
	\centering
	\includegraphics[width=3.5in]{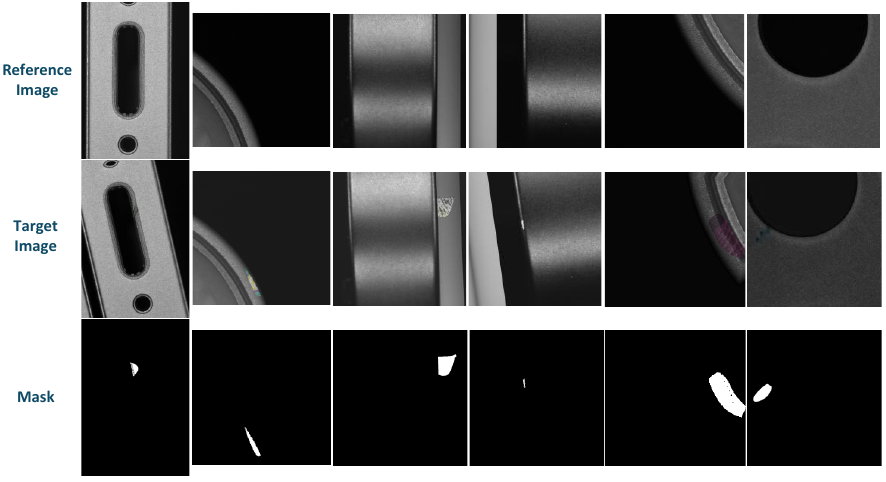} 
	\caption{\Rev{1-1, 2-1}{Examples of synthetic industrial anomaly pairs}}
	\label{fig:synthetic_pairs}
\end{figure}

\Rev{1-1, 2-1}{
\subsection{Synthetic Industrial Anomaly Generation}
\label{sec:synthetic_anomaly_generation}

We construct 5k synthetic industrial image pairs from a private mobile phone back-cover dataset using only defect-free training images, without using any real industrial defect images. The goal is not to faithfully reproduce the full long-tailed distribution of real industrial defects, but to provide controllable reference-target discrepancies for learning difference-based localization.

For each defect-free image, we use it as the normal reference image \(I_{\rm ref}\). With a probability of 0.75, we synthesize an anomalous target by injecting an augmented external texture into an irregular foreground-constrained Perlin region. With a probability of 0.25, we keep the target image defect-free and set the bounding box to \([0,0,0,0]\). The anomaly mask is obtained from thresholded Perlin noise, constrained by the object foreground mask, and reduced to a single connected component to avoid fragmented pseudo-defects. Finally, we apply mild brightness perturbation and small-angle rotation to the target image, while synchronously transforming the mask to preserve accurate localization labels. The ground-truth box is automatically derived as the tight box enclosing the non-zero mask region.

Figure~\ref{fig:synthetic_pairs} provides visual examples of the generated synthetic pairs, including the normal reference image, synthetic target image, anomaly mask, and bounding-box annotation. These synthetic pairs provide localization-oriented supervision for reference-guided difference learning rather than serving as a complete simulation of real industrial defects.
The generation process is summarized in Algorithm~\ref{alg:synthetic_anomaly}.}

\begin{algorithm}[h]
	\caption{\Rev{1-1,2-1}{Synthetic Industrial Image Pair Generation}}
	\label{alg:synthetic_anomaly}
	\begin{algorithmic}[1]
		\Require Normal image $I_{\rm ref}$, texture set $\mathcal{T}$
		\Ensure Target image $I_{\rm tar}$, mask $M$, box $B$
		
		\State Sample $r \sim U(0,1)$
		\If{$r < 0.25$}
		\State $I_{\rm tar} \leftarrow I_{\rm ref}$, $M \leftarrow \mathbf{0}$
		\Else
		\State $I_{\rm tex} \leftarrow$ Augment$(\text{Sample}(\mathcal{T}))$
		\State $M_{\rm p} \leftarrow$ Threshold$(\text{PerlinNoise}())$
		\State $M_{\rm f} \leftarrow$ ForegroundMask$(I_{\rm ref})$
		\State $M \leftarrow M_{\rm p} \cap M_{\rm f}$
		\State Sample $\beta \sim U(0,0.8)$
		\State $I_{\rm tar} \leftarrow I_{\rm ref}\odot(1-M)+((1-\beta)I_{\rm tex}+\beta I_{\rm ref})\odot M$
		\EndIf
		\State $(I_{\rm tar}, M) \leftarrow \text{JitterRotate}(I_{\rm tar}, M)$
		\State $B \leftarrow$ BoundingBox$(M)$
		\State \Return $(I_{\rm ref}, I_{\rm tar}, M, B)$
	\end{algorithmic}
\end{algorithm}

\subsection{AD-DualDiff Dataset}
\begin{figure}[h]
	\centering
	\includegraphics[width=3.5in]{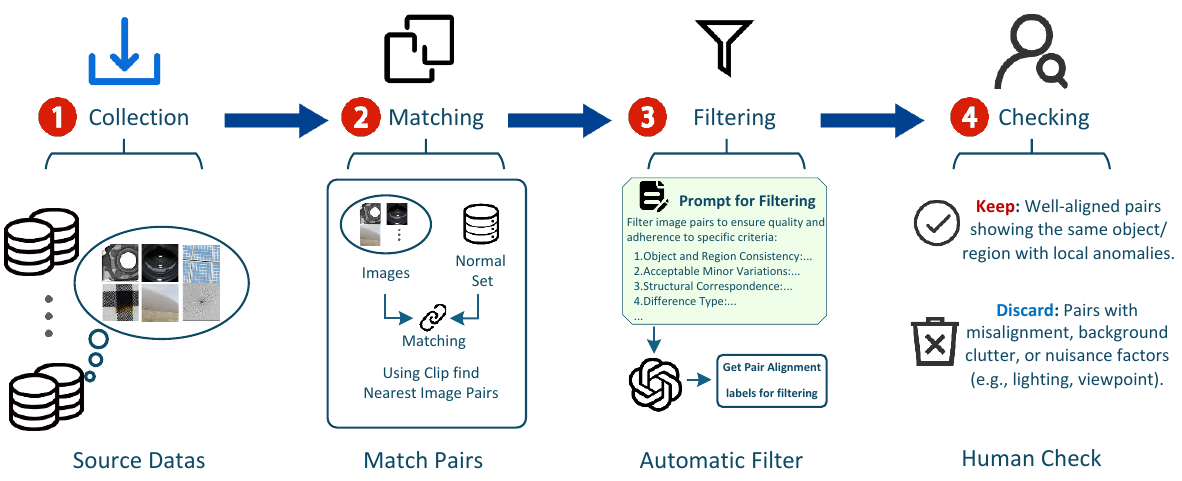} 
	\caption{\Rev{2-5}{
			Construction pipeline of AD-DualDiff. Candidate image pairs are generated by CLIP-based reference matching, filtered by GPT-4o, and manually verified to ensure high-quality reference-target alignment.
	}}
	\label{fig:data_pipeline}
\end{figure}

\begin{figure*}[!t]
	\centering
	\includegraphics[width=\linewidth]{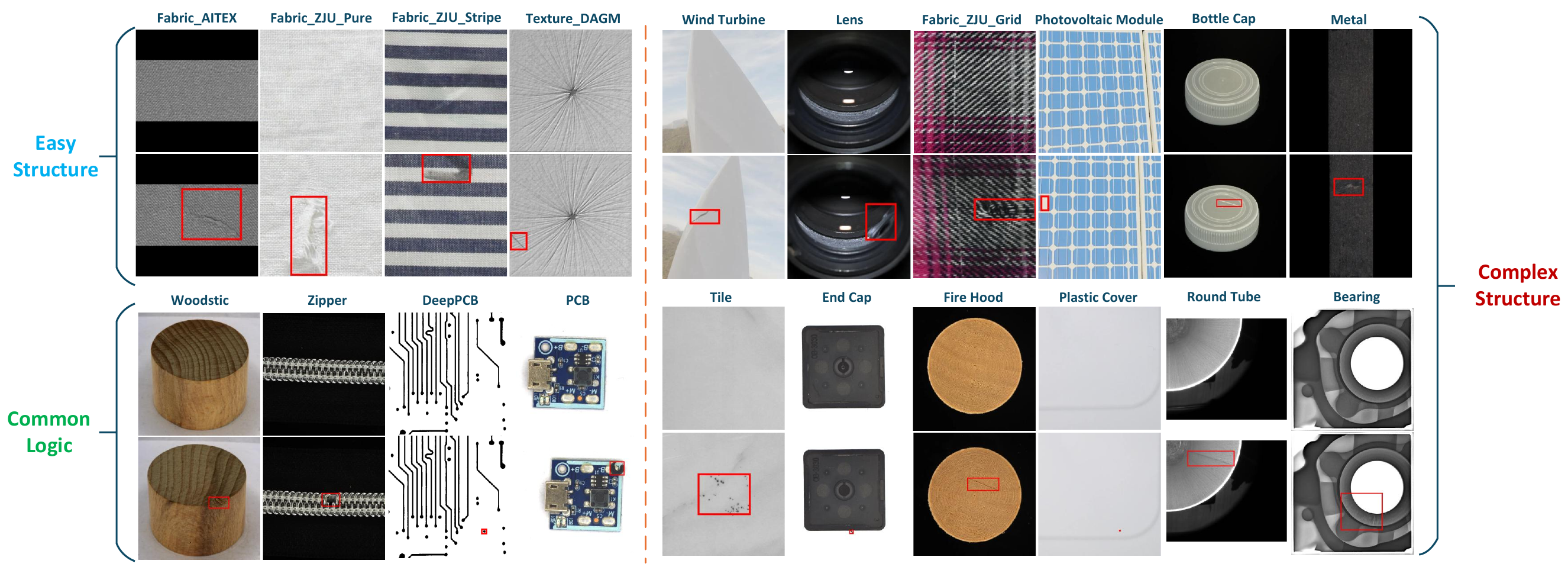} 
	\caption{Visualization of the AD-DualDiff dataset, which is organized into three subsets: Easy structural anomalies, Complex structural anomalies, and Common logical anomalies.}
	\label{fig:dataset_pairs_show}
\end{figure*}

\begin{table}[h]
	\centering
	\caption{Statistical overview of the AD-DualDiff dataset.}
	\label{tab:ad_dualdiff_overview}
	\setlength{\tabcolsep}{3pt}
	\renewcommand{\arraystretch}{1.08}
	\resizebox{\linewidth}{!}{
		\begin{tabular}{l l c c c l}
			\toprule
			\textbf{Subset} &
			\textbf{Category} &
			\textbf{Train} &
			\textbf{Test (Total (N / A))} &
			\textbf{Data Source} &
			\textbf{Link} \\
			\midrule
			
			\multirow{4}{*}{Easy Structure}
			& Fabric\_DAGM & 5,436 & 794 (123 / 671) & DAGM~\cite{DAGM}
			& \href{https://hci.iwr.uni-heidelberg.de/content/weakly-supervised-learning-industrial-optical-inspection}{Project Page} \\
			
			& Fabric\_AITEX & 1,392 & 118 (50 / 68) & AITEX~\cite{AITEX}
			& \href{https://www.aitex.es/afid/}{Project Page} \\
			
			& Fabric\_ZJU\_Pure & 8,513 & 931 (139 / 792) & ZJU~\cite{zju-leaper}
			& \multirow{2}{*}{\href{https://github.com/nico-zck/ZJU-Leaper-Dataset}{GitHub}} \\
			
			& Fabric\_ZJU\_Stripe & 6,534 & 987 (300 / 687) & ZJU~\cite{zju-leaper}
			&  \\
			
			\midrule
			
			\multirow{4}{*}{Common Logic}
			& DeepPCB & 1,497 & 716 (102 / 614) & DeepPCB~\cite{tang2019deeppcb}
			& \href{https://github.com/tangsanli5201/DeepPCB}{GitHub} \\
			
			& Woodstic & 2,386 & 598 (176 / 422) & Real-IAD~\cite{wang2024realiad}
			& \multirow{3}{*}{\href{https://realiad4ad.github.io/Real-IAD/}{Project Page}} \\
			
			& PCB & 2,451 & 1,147 (344 / 803) & Real-IAD~\cite{wang2024realiad}
			&  \\
			
			& Zipper & 2,345 & 474 (174 / 300) & Real-IAD~\cite{wang2024realiad}
			&  \\
			
			\midrule
			
			\multirow{12}{*}{Complex Structure}
			& Fabric\_ZJU\_Grid & 9,285 & 998 (300 / 698) & ZJU~\cite{zju-leaper}
			& \href{https://github.com/nico-zck/ZJU-Leaper-Dataset}{GitHub} \\
			
			& Wind\_Turbine & 7,225 & 797 (168 / 629) & MIAD~\cite{bao2023miad}
			& \multirow{2}{*}{\href{https://miad-2022.github.io/}{Project Page}} \\
			
			& Photovoltaic\_Module & 6,139 & 555 (170 / 385) & MIAD~\cite{bao2023miad}
			&  \\
			
			& Lens & 413 & 327 (106 / 221) & ReinAD~\cite{wangreinad}
			& \multirow{4}{*}{\href{https://huggingface.co/datasets/Tocmac/ReinAD}{HF Dataset}} \\
			
			& Plastic\_Cover & 1,101 & 157 (55 / 102) & ReinAD~\cite{wangreinad}
			&  \\
			
			& Round\_Tube & 313 & 496 (143 / 353) & ReinAD~\cite{wangreinad}
			&  \\
			
			& Bearing & 1,190 & 146 (53 / 93) & ReinAD~\cite{wangreinad}
			&  \\
			
			& Tile & 3,132 & 444 (33 / 411) & Tile~\cite{Tile}
			& \href{https://tianchi.aliyun.com/competition/entrance/531846/information}{Competition} \\
			
			& End\_Cap & 2,381 & 918 (291 / 627) & Real-IAD~\cite{wang2024realiad}
			& \multirow{3}{*}{\href{https://realiad4ad.github.io/Real-IAD/}{Project Page}} \\
			
			& Fire\_Hood & 2,484 & 976 (300 / 676) & Real-IAD~\cite{wang2024realiad}
			&  \\
			
			& Bottle\_Cap & 2,437 & 976 (288 / 688) & Real-IAD~\cite{wang2024realiad}
			&  \\
			
			& Metal & 2,512 & 448 (128 / 320) & KSDD2~\cite{Bozic2021KSDD2}
			& \href{https://www.vicos.si/resources/kolektorsdd2/}{Project Page} \\
			
			\midrule
			
			\textbf{Total} & -- & \textbf{69,166} & \textbf{13,003 (3,443 / 9,560)} & -- & -- \\
			
			\bottomrule
		\end{tabular}
	}
\end{table}
Despite substantial progress in industrial anomaly detection, existing benchmarks are predominantly designed for traditional one-class or multi-class settings and rely almost exclusively on \emph{single-image} inputs.
Representative datasets such as MVTec-AD and VisA typically assume clean and homogeneous backgrounds, limited object categories, and visually salient defect patterns.
In these benchmarks, anomalies are often centrally located and exhibit strong contrast against normal regions, which simplifies recognition but fails to faithfully reflect real-world industrial environments characterized by complex backgrounds and subtle defects.
More importantly, current benchmarks lack explicitly constructed \emph{paired-image} data, making them unsuitable for evaluating difference-based reasoning and reference-guided anomaly localization under dual-image settings.
Recent studies have explored multimodal supervision or large-model-based formulations for industrial anomaly detection.
However, most of these approaches still cast the task as multiple-choice or question-answering problems, rather than grounding anomalies directly within visual content.
\begin{table}[h]
	\centering
	\caption{Comparison with existing industrial anomaly detection benchmarks.}
	\label{tab:benchmark_compare}
	\setlength{\tabcolsep}{3pt}        
	\renewcommand{\arraystretch}{1.1}  
	\resizebox{\linewidth}{!}{
		\begin{tabular}{l c c c c}
			\toprule
			\textbf{Dataset} &
			\textbf{\#Cats} &
			\textbf{\#Images (Train/Test)} &
			\textbf{\#Anomalies} &
			\textbf{Annotation} \\
			\midrule
			
			MVTec AD~\cite{bergmann2019mvtec}
			& 15
			& 5,354 (3,629 / 1,725)
			& 1,258
			& Pixel \\
			
			MVTec LOCO AD
			& 5
			& 3,644 (1,772 / 1,568)
			& 993
			& Pixel \\
			
			VisA~\cite{zou2022visa}
			& 12
			& 10,821 (8,659 / 2,162)
			& 1,200
			& Pixel \\
			
			MulSen-AD~\cite{li2025multi}
			& 15
			& 2,035 (1,391 / 644)
			& 491
			& Pixel \\
			
			Uni-Medical~\cite{darcet2023vision}
			& 3
			& 20,352 (13,339 / 7,013)
			& 4,499
			& Pixel \\
			
			PAD~\cite{zhou2023pad}
			& 20
			& 10,133 (4,629 / 5,504)
			& 4,902
			& Pixel \\
			
			MPDD~\cite{gudovskiy2022mpdd}
			& 6
			& 1,346 (888 / 458)
			& 282
			& Pixel \\
			
			Eyecandies~\cite{bonfiglioli2022eyecandies}
			& 10
			& 15,500 (10,000 / 4,000)
			& 2,250
			& Pixel \\
			
			MIAD~\cite{bao2023miad}
			& 7
			& 105,000 (70,000 / 35,000)
			& 17,500
			& Pixel \\
			
			\midrule
			
			\textbf{AD-DualDiff (Ours)}
			& \textbf{20}
			& \textbf{82,169 (69,166 / 13,003)}
			& \textbf{9,560}
			& \textbf{Pixel \& Box} \\
			
			\bottomrule
		\end{tabular}
	}
\end{table}

To bridge this gap, we introduce \textbf{AD-DualDiff}, a high-quality paired-image industrial anomaly localization dataset.
AD-DualDiff fills a critical void in existing benchmarks by providing carefully curated image pairs that support cross-image comparison and precise anomaly localization.
As shown in Table~\ref{tab:ad_dualdiff_overview}, AD-DualDiff is constructed from multiple public industrial anomaly datasets, including Real-IAD~\cite{wang2024realiad}, ReinAD~\cite{wangreinad}, DeepPCB~\cite{tang2019deeppcb}, DAGM~\cite{DAGM}, KSDD2~\cite{Bozic2021KSDD2}, AITEX~\cite{AITEX}, ZJU-Leaper~\cite{zju-leaper}, MIAD~\cite{bao2023miad}, Tile~\cite{Tile}. Compared with existing industrial anomaly datasets, AD-DualDiff covering a wide range of object categories, acquisition conditions, and defect types, as summarized in Table~\ref{tab:benchmark_compare}, It is worth noting that, to remain consistent with existing anomaly detection benchmarks, AD-DualDiff is also organized into training and test splits. However, the training set is not used for model optimization; instead, it serves solely as a pool of normal reference samples for difference-based comparison. For each test instance, we retrieve the top eight most visually similar defect-free samples using CLIP to provide the corresponding normal references.

	As shown in Fig.~\ref{fig:data_pipeline}, the dataset construction follows a four-stage curation pipeline. \Rev{2-3}{ First, candidate images are manually selected according to object category, anomaly extent, and the availability of paired normal samples, while samples with extremely ambiguous anomalies are excluded. Second, for each anomalous image, CLIP-based retrieval is used to identify the most visually similar normal image from the same category, forming candidate reference-target pairs. Third, GPT-4o~\cite{hurst2024gpt} is adopted for automatic pair-quality filtering, removing more than 380 candidate pairs whose dominant differences are mainly caused by nuisance factors, such as large viewpoint changes, illumination variations, background clutter, or unsuitable reference images. Finally, all remaining pairs are manually reviewed to ensure that the primary visual discrepancy corresponds to the annotated anomaly region rather than background noise, severe viewpoint shifts, or reference mismatch.
	Since different industrial categories vary in candidate sample size, anomaly visibility, and pair-alignment difficulty, the retention ratio is not fixed across categories. We therefore conduct category-adaptive verification while following the same quality criteria, including object/region consistency, acceptable viewpoint and illumination variation, structural correspondence, and consistency between the dominant visual difference and the annotated anomaly region. This combination of automatic filtering and manual verification ensures high-quality paired-image alignment.
}

In contrast to prior benchmarks with relatively homogeneous backgrounds, AD-DualDiff features diverse and cluttered industrial scenes that more closely resemble real production environments.
As shown in the Table~\ref{tab:ad_dualdiff_overview} and Fig.~\ref{fig:dataset_pairs_show}, the dataset includes \emph{easy structural anomalies}, \emph{complex structural anomalies}, and \emph{common logical anomalies}.
\emph{Easy structural anomalies} are characterized by visually salient and well-defined defect patterns that can be readily distinguished from normal structures, as exemplified by the uniformly textured textile images shown in Fig.~\ref{fig:dataset_pairs_show}.
\emph{Complex structural anomalies} involve cluttered or highly textured backgrounds, subtle appearance variations, and small-scale defect regions; as illustrated by the wind turbine and photovoltaic module examples in Fig.~\ref{fig:dataset_pairs_show}, the defects occupy only a small fraction of the image and are not fully aligned, while in the bearing case the defects exhibit no clear boundary from the background, making precise localization significantly more challenging.
\emph{Common logical anomalies} primarily arise from semantic inconsistencies, such as missing components or the presence of foreign objects, which require reasoning beyond low-level visual cues.
This diversity substantially increases the difficulty of localization and reasoning, making AD-DualDiff a rigorous benchmark for evaluating difference-aware anomaly localization models.
For each valid image pair, AD-DualDiff provides aligned bounding box annotations and pixel-level masks for anomalous regions.
These annotations enable systematic evaluation of anomaly localization performance under paired-image settings.
By combining realistic backgrounds, diverse anomaly types, and high-quality paired supervision, AD-DualDiff offers a challenging and comprehensive benchmark for studying difference-driven anomaly localization with large vision-language models.

\section{Experiments}
\begin{figure*}[!t]
	\centering
	\includegraphics[width=\textwidth]{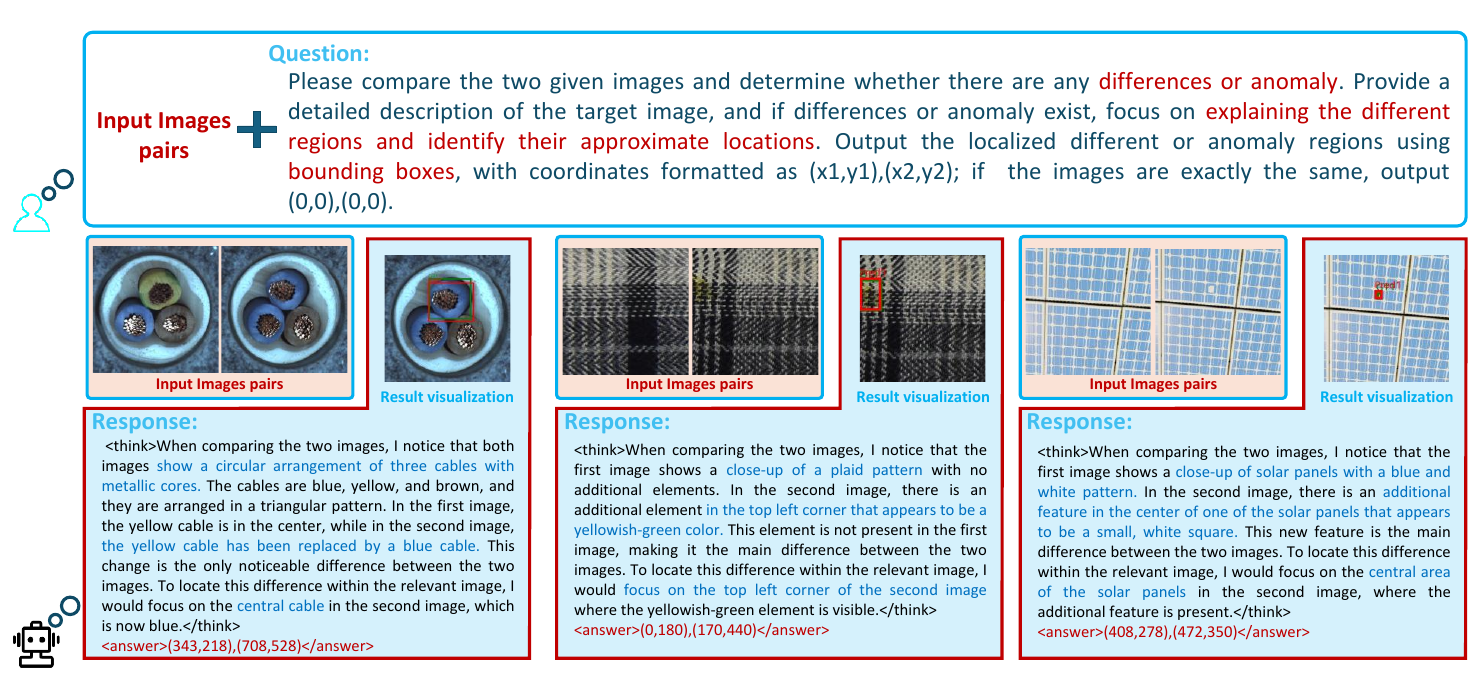} 
	\caption{Illustration of the interpretable difference-aware reasoning process in DifferAD-R1.
	Given paired input images, DifferAD-R1 performs comparative visual reasoning to explicitly explain detected differences in natural language,
	and simultaneously predicts bounding boxes to localize anomalous regions.}
	\label{fig:display_conversation}
\end{figure*}

\begin{figure*}[!t]
	\centering
	\includegraphics[width=\textwidth]{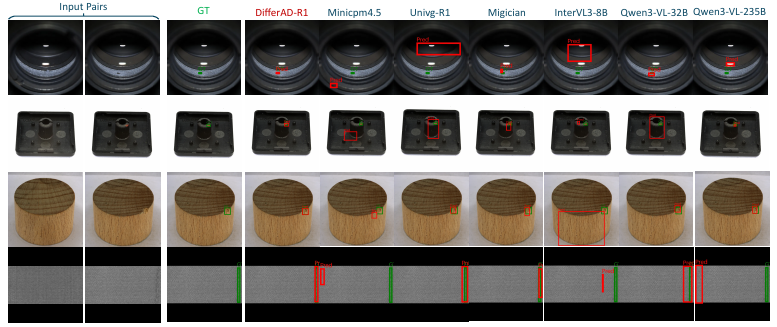} 
	\caption{Qualitative comparison of visual localization performance across different models.
	The figure compares anomaly localization results produced by multiple methods on paired-image inputs,
	covering a wide range of anomaly types, including subtle defects, small stains, missing regions, and large boundary anomalies.
	Compared with baseline models, DifferAD-R1 yields more precise and consistent bounding boxes,
	demonstrating superior visual grounding and localization capability.}

	\label{fig:display_photo}
\end{figure*}

\begin{table}[h]
	\caption{Performance comparison on the proposed AD-DualDiff dataset.}
	\label{tab:output_dualdiff}
	\centering
	\resizebox{\linewidth}{!}{
		\begin{tabular}{lcccc}
			\hline
			Method & Acc@0.3 (\%) & F1 (\%) & mIoU (\%) & $mIoU_{ano}$ (\%) \\
			\hline
			Mantis\cite{jiang2024mantis} & 27.76 & 0.00 & 27.76 & 27.76 \\
			Minicpm2.6\cite{yao2024minicpm} & 20.20 & 3.96 & 20.30 & 2.38 \\
			Minicpm4.5\cite{yu2025minicpm4.5} & 6.08 & 9.21 & 5.39 & 6.03 \\
			mPLUG-Owl3\cite{ye2024mplug} & 5.51 & 9.54 & 4.72 & 6.44 \\
			Migician\cite{li2025migician} & 37.26 & 40.81 & 28.85 & 23.81 \\
			UniVG-R1\cite{bai2025univg} & 32.20 & 43.16 & 23.51 & 30.06 \\
			InternVL2-8B\cite{cai2024internlm2} & 23.49 & 2.36 & 23.34 & 1.10 \\
			InternVL3-8B\cite{zhu2025internvl3} & 28.36 & 10.10 & 27.50 & 4.90 \\
			InternVL3.5-8B\cite{wang2025internvl35} & 7.51 & 7.09 & 7.36 & 4.87 \\
			Qwen2VL-7B\cite{wang2024qwen2vl} & 41.25 & 26.06 & 36.82 & 12.28 \\
			Qwen2.5VL-7B\cite{bai2025qwen25vl} & 22.58 & 4.62 & 22.64 & 3.42 \\
			Qwen3VL-8B\cite{Bai2025Qwen3VLTR} & 39.63 & 28.30 & 35.53 & 13.35 \\
			Qwen3VL-32B\cite{Bai2025Qwen3VLTR} & 45.49 & 41.16 & 38.21 & 21.36 \\
			Qwen3VL-235B\cite{Bai2025Qwen3VLTR} & 49.48 & 44.57 & 41.33 & 21.61 \\
			\hline
			DifferAD-R1 (Ours) & \textbf{61.12} & \textbf{60.95} & \textbf{49.21} & \textbf{37.66} \\
			\hline
		\end{tabular}
	}
\end{table}

\begin{table}[h]
	\caption{Performance comparison on the MVTec AD.}
	\label{tab:output_MvTecAD}
	\centering
	\resizebox{\linewidth}{!}{
		\begin{tabular}{lcccc}
			\hline
			Method & Acc@0.3 (\%) & F1 (\%) & mIoU (\%) & $mIoU_{ano}$ (\%) \\
			\hline
			Mantis\cite{jiang2024mantis} & 27.70 & 0.00 & 27.70 & 0.00 \\
			Minicpm2.6\cite{yao2024minicpm} & 20.87 & 9.63 & 20.26 & 4.39 \\
			Minicpm4.5\cite{yu2025minicpm4.5} & 14.69 & 24.39 & 10.75 & 15.06 \\
			mPLUG-Owl3\cite{ye2024mplug} & 14.11 & 23.71 & 11.71 & 16.45 \\
			Migician\cite{li2025migician} & 48.76 & 57.92 & 37.53 & 36.90 \\
			UniVG-R1\cite{bai2025univg} & 36.32 & 51.28 & 26.99 & 36.92 \\
			InternVL2-8B\cite{cai2024internlm2} & 26.96 & 8.39 & 26.20 & 3.84 \\
			InternVL3-8B\cite{zhu2025internvl3} & 37.49 & 24.92 & 34.62 & 12.17 \\
			InternVL3.5-8B\cite{wang2025internvl35} & 13.32 & 16.35 & 11.88 & 10.61 \\
			Qwen2VL-7B\cite{wang2024qwen2vl} & 43.84 & 32.61 & 38.90 & 15.72 \\
			Qwen2.5VL-7B\cite{bai2025qwen25vl} & 28.07 & 15.08 & 26.41 & 8.04 \\
			Qwen3VL-8B\cite{Bai2025Qwen3VLTR} & 56.57 & 57.99 & 45.64 & 31.38 \\
			Qwen3VL-32B\cite{Bai2025Qwen3VLTR} & 56.63 & 61.44 & 45.30 & 35.69 \\
			Qwen3VL-235B\cite{Bai2025Qwen3VLTR} & \textbf{70.93} & 73.09 & \textbf{57.14} & 42.38 \\
			\hline
			DifferAD-R1 (Ours) & 70.40 & \textbf{74.68} & 56.21 & \textbf{49.21} \\
			\hline
		\end{tabular}
	}
\end{table}


\begin{table*}[t]
	\caption{Per-class localization comparison on MVTec-AD and AD-DualDiff (metrics in \%). Each entry is \textbf{(mIoU, Acc@0.3, F1@0.3)}. Missing results are left as ``--''.}
	\label{tab:per_class_mvtec_ad_dualdiff}
	\centering
	\setlength{\tabcolsep}{4pt}
	\renewcommand{\arraystretch}{1.08}
	\resizebox{\textwidth}{!}{
		\begin{tabular}{l l l c c c c c}
			\toprule
			\textbf{Dataset} & \textbf{Type} & \textbf{Class} &
			\textbf{UniVG-R1} & \textbf{Migician} & \textbf{Qwen3VL-32B} & \textbf{Qwen3VL-235B} & \textbf{DifferAD-R1 (Ours)} \\
			\midrule
			
			\multirow{17}{*}{\textbf{MVTec-AD}} &
			\multirow{6}{*}{Texture} &
			carpet      & (35.40, 58.12, 73.51) & (50.45, 70.94, 77.03) & (50.36, 64.96, 70.50) & (\textcolor{red}{62.34}, \textcolor{blue}{83.76}, \textcolor{blue}{88.05}) & (\textcolor{blue}{62.31}, \textcolor{red}{86.32}, \textcolor{red}{90.48}) \\
			& & grid        & (25.85, 34.62, 51.43) & (48.70, 57.69, 61.18) & (43.59, 50.00, 51.85) & (\textcolor{blue}{58.33}, \textcolor{blue}{76.92}, \textcolor{blue}{81.63}) & (\textcolor{red}{68.96}, \textcolor{red}{89.74}, \textcolor{red}{92.59}) \\
			& & leather     & (28.24, 45.97, 62.98) & (44.42, 56.45, 59.70) & (47.09, 66.94, 72.11) & (\textcolor{red}{62.45}, \textcolor{red}{85.48}, \textcolor{red}{89.16}) & (\textcolor{blue}{56.16}, \textcolor{blue}{77.42}, \textcolor{blue}{82.93}) \\
			& & tile        & (52.78, 61.54, 76.19) & (72.76, 85.47, 89.70) & (72.95, 84.62, 88.16) & (\textcolor{blue}{81.68}, \textcolor{red}{94.02}, \textcolor{red}{95.65}) & (\textcolor{red}{82.40}, \textcolor{blue}{90.60}, \textcolor{blue}{93.41}) \\
			& & wood        & (19.70, 21.52, 35.42) & (40.87, 48.10, \textcolor{blue}{56.84}) & (33.76, 37.97, 46.15) & (\textcolor{blue}{47.03}, \textcolor{blue}{49.37}, 50.00) & (\textcolor{red}{73.08}, \textcolor{red}{86.08}, \textcolor{red}{91.47}) \\
			& & \textbf{Texture (Avg.)}
			& (32.39, 44.35, 59.91)
			& (51.44, 63.73, 68.89)
			& (49.55, 60.90, 65.75)
			& (\textcolor{blue}{62.37}, \textcolor{blue}{77.91}, \textcolor{blue}{80.90})
			& (\textcolor{red}{68.58}, \textcolor{red}{86.03}, \textcolor{red}{90.18}) \\
			
			\cmidrule(lr){2-8}
			& \multirow{11}{*}{Object} &
			bottle      & (36.76, 46.99, 63.93) & (47.89, 65.06, 77.52) & (47.87, 59.04, 66.00) & (\textcolor{red}{69.44}, \textcolor{red}{85.54}, \textcolor{red}{89.47}) & (\textcolor{blue}{66.49}, \textcolor{blue}{81.93}, \textcolor{blue}{87.40}) \\
			& & cable       & (20.29, 29.33, 45.60) & (25.80, 39.33, 56.46) & (39.03, 50.00, 58.10) & (\textcolor{red}{63.95}, \textcolor{red}{78.00}, \textcolor{red}{79.75}) & (\textcolor{blue}{47.48}, \textcolor{blue}{60.00}, \textcolor{blue}{65.52}) \\
			& & capsule     & (16.80, 27.27, 42.86) & (28.62, 40.15, 51.53) & (32.74, 42.42, 48.65) & (\textcolor{blue}{36.64}, \textcolor{blue}{50.76}, \textcolor{blue}{58.60}) & (\textcolor{red}{36.95}, \textcolor{red}{54.55}, \textcolor{red}{65.12}) \\
			& & hazelnut    & (38.25, 51.82, 68.26) & (38.46, 57.27, 71.86) & (37.71, 55.45, 70.66) & (\textcolor{blue}{65.30}, \textcolor{blue}{80.00}, \textcolor{blue}{84.06}) & (\textcolor{red}{69.30}, \textcolor{red}{86.36}, \textcolor{red}{89.93}) \\
			& & metal\_nut  & (38.80, 44.35, 61.45) & (49.21, 60.87, 72.05) & (\textcolor{blue}{66.19}, \textcolor{blue}{81.74}, \textcolor{blue}{87.72}) & (\textcolor{red}{69.02}, \textcolor{red}{87.83}, \textcolor{red}{91.86}) & (58.79, 72.17, 80.25) \\
			& & pill        & (33.99, 46.11, 63.11) & (39.09, 55.69, 68.38) & (38.55, 53.29, 64.55) & (\textcolor{blue}{46.32}, \textcolor{blue}{62.87}, \textcolor{blue}{71.82}) & (\textcolor{red}{58.48}, \textcolor{red}{74.85}, \textcolor{red}{82.50}) \\
			& & screw       & (10.93, 15.00, 23.60) & ( 8.15, 10.63, 18.29) & (\textcolor{red}{35.33}, \textcolor{red}{43.13}, \textcolor{red}{40.52}) & (\textcolor{blue}{34.87}, \textcolor{red}{43.13}, 38.10) & (33.88, 42.50, \textcolor{blue}{40.26}) \\
			& & toothbrush  & (24.72, 30.95, 47.27) & (22.78, 30.95, 47.27) & (47.66, \textcolor{blue}{59.52}, \textcolor{red}{65.31}) & (\textcolor{blue}{51.16}, \textcolor{red}{61.90}, \textcolor{blue}{63.64}) & (\textcolor{red}{51.22}, 57.14, 60.87) \\
			& & transistor  & ( 7.42, 12.00, 21.43) & (25.98, 31.00, 28.87) & (\textcolor{blue}{53.36}, \textcolor{blue}{62.00}, \textcolor{blue}{51.28}) & (\textcolor{red}{72.18}, \textcolor{red}{80.00}, \textcolor{red}{67.74}) & (33.07, 38.00, 31.11) \\
			& & zipper      & (14.91, 19.21, 32.22) & (19.79, 21.85, 32.18) & (32.54, 38.41, 40.00) & (\textcolor{blue}{36.45}, \textcolor{blue}{44.37}, \textcolor{blue}{46.84}) & (\textcolor{red}{44.60}, \textcolor{red}{58.28}, \textcolor{red}{66.31}) \\
			& & \textbf{Object (Avg.)}
			& (24.29, 32.30, 46.97)
			& (30.58, 41.28, 52.44)
			& (43.10, 54.50, 59.28)
			& (\textcolor{red}{54.53}, \textcolor{red}{67.44}, \textcolor{red}{69.19})
			& (\textcolor{blue}{50.03}, \textcolor{blue}{62.58}, \textcolor{blue}{66.93}) \\
			
			\cmidrule(lr){1-8}
			\textbf{MVTec-AD} & \textbf{ALL (Avg.)} & \textbf{--} &
			(26.99, 36.32, 51.28) &
			(37.53, 48.76, 57.92) &
			(45.25, 56.63, 61.44) &
			(\textcolor{red}{57.14}, \textcolor{red}{70.93}, \textcolor{blue}{73.09}) &
			(\textcolor{blue}{56.21}, \textcolor{blue}{70.40}, \textcolor{red}{74.68}) \\
			
			\midrule
			
			\multirow{23}{*}{\textbf{AD-DualDiff}} &
			\multirow{5}{*}{Easy\_structure} &
			Texture\_DAGM & (\textcolor{blue}{50.76}, \textcolor{blue}{72.17}, \textcolor{blue}{83.81}) & (49.76, 72.29, 80.60) & (28.38, 30.35, 29.91) & (36.81, 48.11, 55.70) & (\textcolor{red}{70.66}, \textcolor{red}{93.95}, \textcolor{red}{96.31}) \\
			& & Fabric\_AITEX & (25.99, 35.59, \textcolor{blue}{47.22}) & (49.13, 53.39, 44.44) & (48.59, 54.24, 42.55) & (\textcolor{blue}{50.47}, \textcolor{blue}{55.93}, 40.91) & (\textcolor{red}{61.05}, \textcolor{red}{70.34}, \textcolor{red}{65.35}) \\
			& & Fabric\_ZJU\_Pure & (47.95, 65.84, 79.38) & (48.97, 64.98, 75.45) & (\textcolor{blue}{50.28}, \textcolor{blue}{70.78}, \textcolor{blue}{80.40}) & (45.89, 62.19, 72.33) & (\textcolor{red}{59.65}, \textcolor{red}{80.77}, \textcolor{red}{87.68}) \\
			& & Fabric\_ZJU\_Stripe & (45.36, 54.51, 70.28) & (44.49, 61.09, 74.93) & (68.26, 85.82, 88.98) & (\textcolor{blue}{71.32}, \textcolor{blue}{89.77}, \textcolor{blue}{92.12}) & (\textcolor{red}{74.79}, \textcolor{red}{89.97}, \textcolor{red}{92.67}) \\
			& & \textbf{Easy\_structure (Avg.)}
			& (42.52, 57.03, 70.17)
			& (48.09, 62.94, 68.86)
			& (48.88, 60.30, 60.46)
			& (\textcolor{blue}{51.12}, \textcolor{blue}{64.00}, \textcolor{blue}{65.27})
			& (\textcolor{red}{66.54}, \textcolor{red}{83.76}, \textcolor{red}{85.50}) \\
			
			\cmidrule(lr){2-8}
			& \multirow{5}{*}{Common\_logic} &
			DeepPCB & (19.36, 37.43, 54.47) & (\textcolor{blue}{25.28}, \textcolor{blue}{44.69}, \textcolor{blue}{61.48}) & (19.17, 23.46, 29.56) & (14.80, 15.36, 13.92) & (\textcolor{red}{36.53}, \textcolor{red}{63.41}, \textcolor{red}{76.44}) \\
			& & PCB & ( 6.07,  9.33, 17.07) & ( 7.83, 11.86, 20.95) & (\textcolor{blue}{35.82}, \textcolor{blue}{37.93}, 22.61) & (\textcolor{red}{42.66}, \textcolor{red}{49.61}, \textcolor{red}{43.77}) & (24.83, 30.34, \textcolor{blue}{31.44}) \\
			& & Woodstic & ( 8.69, 11.87, 21.23) & (19.98, 24.08, 24.33) & (25.30, 28.93, 23.42) & (\textcolor{red}{43.54}, \textcolor{red}{53.68}, \textcolor{red}{54.06}) & (\textcolor{blue}{42.59}, \textcolor{blue}{53.01}, \textcolor{blue}{53.86}) \\
			& & Zipper & ( 8.88, 12.45, 22.14) & ( 6.64,  7.17, 13.04) & (\textcolor{red}{43.94}, \textcolor{blue}{46.84}, \textcolor{blue}{30.77}) & (43.52, 45.78, 25.07) & (\textcolor{blue}{43.89}, \textcolor{red}{47.05}, \textcolor{red}{32.75}) \\
			& & \textbf{Common\_logic (Avg.)}
			& (10.75, 17.77, 28.73)
			& (14.93, 21.95, 29.95)
			& (31.06, 34.29, 26.59)
			& (\textcolor{blue}{36.13}, \textcolor{blue}{41.11}, \textcolor{blue}{34.20})
			& (\textcolor{red}{36.96}, \textcolor{red}{48.45}, \textcolor{red}{48.62}) \\
			
			\cmidrule(lr){2-8}
			& \multirow{13}{*}{Complex\_structure} &
			Fabric\_ZJU\_Grid & (42.19, 54.01, 69.94) & (55.87, 73.35, \textcolor{blue}{81.00}) & (56.69, 74.05, 79.75) & (\textcolor{blue}{58.37}, \textcolor{blue}{75.35}, 79.53) & (\textcolor{red}{66.62}, \textcolor{red}{84.67}, \textcolor{red}{88.69}) \\
			& & Wind\_Turbine & (\textcolor{blue}{52.27}, \textcolor{blue}{66.50}, \textcolor{blue}{79.73}) & (34.39, 49.94, 63.35) & (40.89, 56.84, 63.56) & (36.20, 47.55, 52.06) & (\textcolor{red}{66.89}, \textcolor{red}{81.18}, \textcolor{red}{87.84}) \\
			& & Tile & (\textcolor{blue}{52.20}, \textcolor{blue}{74.77}, \textcolor{blue}{85.49}) & (36.17, 56.31, 71.13) & (35.48, 52.03, 66.24) & (30.79, 41.89, 55.36) & (\textcolor{red}{56.65}, \textcolor{red}{77.03}, \textcolor{red}{86.29}) \\
			& & Bearing & (23.30, 31.51, 43.18) & (46.10, 53.42, 46.03) & (\textcolor{blue}{51.41}, \textcolor{blue}{60.96}, \textcolor{blue}{55.81}) & (44.27, 47.95, 30.91) & (\textcolor{red}{54.83}, \textcolor{red}{64.38}, \textcolor{red}{61.76}) \\
			& & Bottle\_cap & (13.29, 18.03, 29.33) & (27.64, 30.33, 29.46) & (47.63, 57.99, 57.73) & (\textcolor{red}{55.04}, \textcolor{red}{72.64}, \textcolor{red}{76.04}) & (\textcolor{blue}{48.02}, \textcolor{blue}{59.73}, \textcolor{blue}{61.36}) \\
			& & End\_cap & ( 4.02,  4.58,  8.75) & (17.13, 15.80,  6.98) & (33.40, 35.40, 20.43) & (\textcolor{red}{39.27}, \textcolor{red}{44.99}, \textcolor{red}{37.42}) & (\textcolor{blue}{38.19}, \textcolor{blue}{42.12}, \textcolor{blue}{36.74}) \\
			& & Fire\_Hood & (11.40, 15.68, 26.97) & (28.48, 29.51, 20.92) & (30.69, 33.71, 24.68) & (\textcolor{red}{49.61}, \textcolor{red}{60.35}, \textcolor{red}{60.55}) & (\textcolor{blue}{42.73}, \textcolor{blue}{47.75}, \textcolor{blue}{41.24}) \\
			& & Photovoltaic\_Module & ( 3.49,  5.23, \textcolor{blue}{ 9.00}) & (10.92, 11.71,  5.41) & (30.37, 29.55,  2.98) & (\textcolor{blue}{32.92}, \textcolor{blue}{32.43},  5.06) & (\textcolor{red}{35.25}, \textcolor{red}{40.72}, \textcolor{red}{34.33}) \\
			& & Lens & (10.09, 14.68, 25.20) & (18.39, 24.16, \textcolor{blue}{28.82}) & (24.48, 26.91, 21.12) & (\textcolor{blue}{28.49}, \textcolor{blue}{31.80}, 18.91) & (\textcolor{red}{33.04}, \textcolor{red}{39.45}, \textcolor{red}{39.26}) \\
			& & Metal & (30.73, 42.86, 60.00) & (29.59, 42.63, 56.95) & (\textcolor{blue}{45.99}, \textcolor{blue}{60.27}, \textcolor{blue}{65.64}) & (41.63, 51.34, 51.98) & (\textcolor{red}{67.23}, \textcolor{red}{85.49}, \textcolor{red}{89.22}) \\
			& & Plastic\_Cover & ( 6.32, 10.83, 18.60) & (11.44, 11.47,  4.14) & (32.36, 31.85, \textcolor{blue}{18.91}) & (\textcolor{blue}{37.04}, \textcolor{blue}{38.22}, 15.65) & (\textcolor{red}{37.38}, \textcolor{red}{46.50}, \textcolor{red}{43.24}) \\
			& & Round\_Tube & ( 7.80,  6.05, \textcolor{blue}{11.41}) & ( 8.81,  7.06,  6.87) & (14.97, 12.10,  8.40) & (\textcolor{red}{23.96}, \textcolor{red}{24.60}, 10.10) & (\textcolor{blue}{23.24}, \textcolor{red}{24.60}, \textcolor{red}{12.62}) \\
			& & \textbf{Complex\_structure (Avg.)}
			& (21.43, 28.73, 38.97)
			& (27.08, 33.81, 35.09)
			& (37.03, 44.30, 40.44)
			& (\textcolor{blue}{39.80}, \textcolor{blue}{47.43}, \textcolor{blue}{41.13})
			& (\textcolor{red}{47.51}, \textcolor{red}{57.80}, \textcolor{red}{56.88}) \\
			
			\cmidrule(lr){1-8}
			\textbf{AD-DualDiff} & \textbf{ALL (Avg.)} & \textbf{--} &
			(23.51, 32.20, 43.16) &
			(28.85, 37.26, 40.82) &
			(38.21, 45.50, 41.67) &
			(\textcolor{blue}{41.33}, \textcolor{blue}{49.48}, \textcolor{blue}{44.57}) &
			(\textcolor{red}{49.21}, \textcolor{red}{61.12}, \textcolor{red}{60.95}) \\
			
			\bottomrule
		\end{tabular}
	}
\end{table*}

\subsection{Experimental Setup}
\paragraph{Datasets}

DifferAD-R1 is initialized from UniVG-R1, which has been aligned on multi-image reasoning tasks in natural scenes.
To further enhance cross-image difference perception, we introduce paired data from natural scenes and synthetic industrial anomalies during the second training stage.
Specifically, the training set for the RL stage consists of 5k COCO-Inpainted~\cite{yan2025coco} natural-scene image pairs and 5k synthetic industrial anomaly pairs, resulting in a total of 10k training samples.
\Rev{2-1}{
The synthetic industrial anomaly pairs are generated from defect-free mobile phone backplate images, which are not included in either AD-DualDiff or MVTec AD.
Therefore, the industrial images used for synthetic anomaly generation have no image-level, category-level, or source-level overlap with the evaluation datasets.
Moreover, no real industrial defect images are used during training; all real anomalous samples are reserved exclusively for evaluation.}
Synthetic anomalies is designed to provide controllable discrepancy-based localization supervision rather than to reproduce the full distribution of real industrial defects.
Therefore, the task is formulated as a general anomaly localization problem in a one-shot setting.
This paradigm is adopted to verify that the anomaly localization capability of DifferAD-R1 generalizes from natural scenes and synthetic phone-backplate anomalies to real-world industrial scenarios from different domains.
\paragraph{Evaluation Details}
We evaluate DifferAD-R1 on the proposed AD-DualDiff dataset as well as the MVTec-AD benchmarks,
which together simulate diverse and challenging real-world industrial inspection environments.
For anomalous samples, ground-truth bounding boxes are obtained by extracting the minimum enclosing rectangles from pixel-level anomaly masks.
For normal samples, the ground-truth bounding box is uniformly set to $(0,0,0,0)$.
Predicted bounding boxes are parsed from the structured output format \texttt{<answer>}$\cdots$\texttt{</answer>} in DifferAD-R1.
We conduct systematic comparisons with representative open-source methods that support anomaly localization
to comprehensively validate the effectiveness of the proposed approach.

\Rev{1-4,2-4}{
	For fair and reproducible comparison, all MLLM baselines supporting multi-image input are evaluated with the same reference-target image pair and a unified dual-image comparison instruction.
	The instruction asks the model to compare the two images and localize the different or anomalous region in the target image.
	Since different MLLMs use different native output formats, we only adapt the format constraint to each model while keeping the same semantic task instruction.
	All predicted boxes are extracted using the same parser; invalid or non-parsable outputs are assigned an IoU of 0, and multiple-box outputs are handled by keeping the first valid box under the single-anomaly setting. The detailed prompts and bounding-box extraction rules are available in our released code repository: \url{https://github.com/Rong2026/work-1/blob/main/DifferAD-R1/baseline_comparison_protocol.md}.
}

\Rev{1-5,2-4}{
	For traditional non-LLM baselines, including PaDiM, PatchCore, and WinCLIP, we follow their standard protocols on AD-DualDiff. PaDiM and PatchCore are trained only on normal training images without defect labels, while WinCLIP uses normal reference images under its few-shot setting. Since these methods output anomaly maps, we convert them into bounding boxes for localization evaluation. Specifically, 30\% of the evaluation data is used to calibrate the threshold that maximizes the F1-score. Pixels above the threshold are treated as anomalous, and the largest connected component is converted into a bounding box for computing mIoU, Acc@0.3, and F1-score under the single-anomaly setting.
}

To comprehensively assess localization accuracy and the ability to distinguish between normal and anomalous samples,
while accounting for the small scale and high difficulty of industrial defects,
we adopt Acc@0.3 to evaluate localization accuracy,
F1-score to measure classification precision and recall,
and $mIoU_{ano}$ and $mIoU_{nor}$ to evaluate localization performance on anomalous and normal samples, respectively.
Here, $mIoU_{ano}$ denotes the average IoU between predicted and ground-truth bounding boxes for anomalous samples,
For a normal sample, a correct prediction of an empty box $(0,0,0,0)$ is assigned an IoU score of 1;
otherwise, the IoU score is set to 0.

\paragraph{Implementation Details}
By default, DifferAD-R1 is trained on top of UniVG-R1, whose vision-language backbone is Qwen2-VL-7B.
The training process is implemented using DeepSpeed to enable efficient and stable optimization of large-scale models.
The learning rate is set to $1\times10^{-6}$ with an effective batch size of 16,
and the visual encoder is frozen to stabilize multimodal alignment.
During GRPO training, 8 responses are sampled per input,
with a temperature of 0.9 and a maximum generation length of 256 tokens. In DCLR, the weight $\alpha$ is set to 0.6, $\lambda$ is set to 0.3, and the IoU threshold $\tau$ is set to 0.3. All experiments are conducted on $4\times A100 GPUs$.

\subsection{Main Results}

We evaluate {DifferAD-R1 against a diverse set of open-source vision-language models with localization capability on both the proposed AD-DualDiff dataset and the MVTec-AD benchmark.
The comparison includes representative models from multiple families, such as Mantis\cite{jiang2024mantis}, MiniCPM\cite{yao2024minicpm}, mPLUG-Owl3\cite{ye2024mplug}, Migician\cite{li2025migician}, UniVG-R1\cite{bai2025univg}, InternVL\cite{cai2024internlm2}, and Qwen-VL\cite{wang2024qwen2vl}, covering a wide range of model scales.
As shown in Tables~\ref{tab:output_dualdiff} and~\ref{tab:output_MvTecAD},
DifferAD-R1 consistently achieves the best overall performance on both datasets,
obtaining \textbf{61.12\% Acc@0.3 / 60.95\% F1} on AD-DualDiff and
\textbf{70.40\% Acc@0.3 / 74.68\% F1} on MVTec-AD.
Notably, despite being trained with a relatively smaller backbone,
DifferAD-R1 outperforms several significantly larger models, including 32B-scale VLMs,
highlighting the effectiveness of paired-image difference modeling and reinforcement learning--based alignment for industrial anomaly localization.
Moreover, these results are achieved without using real industrial defect samples during training,
demonstrating strong cross-dataset generalization.
As illustrated in Table~\ref{tab:per_class_mvtec_ad_dualdiff}, we present a detailed comparison on the MVTec-AD and AD-DualDiff datasets. The proposed method not only significantly outperforms representative baselines such as UniVG-R1 and Migician, but also achieves competitive performance against the much larger Qwen3-VL-235B model.
In particular, on the AD-DualDiff dataset across the three anomaly categories, for the F1-score, our method surpasses Qwen3-VL-235B by 20.23\% on the Easy-structure subset, 14.42\% on the Common-logic subset, and 15.75\% on the Complex-structure subset, respectively.
These results demonstrate the effectiveness of the proposed approach in handling complex industrial anomaly localization tasks.

Furthermore, our difference comparison learning paradigm demonstrates greater robustness in leveraging reference information to improve performance. By first describing normal images followed by the images to be detected, this paradigm compares their differences to accurately reflect the locations of defects. To provide more intuitive analytical evidence, we visualize the detection results of DifferAD-R1 under easy structural anomalies, complex structural anomalies, and common logical anomalies in Fig.~\ref{fig:display_conversation}.

\begin{table}[t]
	\begin{reviewbluetable}{1-5,3-3}
	\caption{Comparison with traditional non-LLM anomaly detection methods on AD-DualDiff.}
	\label{tab:traditional_baselines_addualdiff}
	\centering
	\resizebox{\linewidth}{!}{
		\begin{tabular}{lcccc}
			\hline
			Method & Training Data & mIOU & Acc@0.3 (\%) & F1-score (\%) \\
			\hline
			PaDiM~\cite{defard2021padim} 
			& Normal images only & 33.69 & 37.21 & 22.67 \\
			PatchCore~\cite{roth2022patchcore} 
			& Normal images only & 41.38 & 51.29 & 47.76 \\
			WinCLIP~\cite{jeong2023winclip} 
			& Normal reference images & 30.75 & 33.89 & 17.76 \\
			\hline
			DifferAD-R1 Full (Ours) 
			& Synthetic pairs only & \textbf{49.21} & \textbf{61.12} & \textbf{60.95} \\
			\hline
		\end{tabular}
	}
	\end{reviewbluetable}
\end{table}

Fig.~\ref{fig:display_photo} presents the qualitative comparison results of the DifferAD-R1 anomaly localization model on the AD benchmark dataset. Defective objects are in various real-world scenarios, including complex backgrounds, cross-scale sizes, and blurred edges, which pose challenges to models regarding texture differences, weak edges, and varying target sizes. All results are compared and evaluated using the proposed DifferAD-R1 framework and those of other Multimodal Large Language Models (MLLMs). The results show that existing MLLM methods perform poorly when dealing with small targets or those with blurred edges, often generating overly large or inaccurate bounding boxes (as shown in Fig.~\ref{fig:display_photo}). Some methods suffer from hallucination, tend to over-detect normal regions, and easily misclassify reflective non-defective regions as anomalies, as illustrated in the first row of Fig.~\ref{fig:display_photo}. In contrast, under the guidance of normal comparison, DifferAD-R1 can effectively capture the difference information between the test image and normal images, achieving accurate and stable detection for cross-scale and blurred-edge targets, demonstrating its robustness and generalization ability in various complex industrial scenarios.

\Rev{1-5}{
	Table~\ref{tab:traditional_baselines_addualdiff} further compares DifferAD-R1 with representative traditional non-LLM anomaly detection methods on AD-DualDiff. PaDiM and PatchCore are trained using only normal images from the training split, while WinCLIP uses normal reference images following its standard setting. Since these methods produce anomaly maps rather than bounding boxes, we convert the anomaly maps into boxes using the unified protocol. As shown in the table~\ref{tab:traditional_baselines_addualdiff}, DifferAD-R1 consistently outperforms PaDiM, PatchCore, and WinCLIP across all metrics. These results show that DifferAD-R1 achieves superior performance on AD-DualDiff compared with traditional non-LLM baselines, especially in a more diverse paired-image setting with multiple product categories, varying normal-reference appearances, and relatively small anomalous regions., the results suggest that our reference-guided comparison framework is advantageous for challenging paired-image anomaly localization scenarios.
}

\subsection{Ablation Results}

\begin{table}[h]
	\begin{reviewbluetable}{1-2,3-2,3-3}
		\caption{Ablation study on the core design choices of DifferAD-R1. 
			\cmark/\xmark indicates whether the corresponding component is enabled.}
		\label{tab:ablation_paradigm}
		\centering
		\resizebox{\linewidth}{!}{
			\begin{tabular}{lccc|cc|cc}
				\hline
				\multirow{2}{*}{Variant} &
				\multirow{2}{*}{\makecell{$P_{\rm dual}$\\Dual-image}} &
				\multirow{2}{*}{\makecell{$C_{\rm cold}$\\Dual-Cold-start}} &
				\multirow{2}{*}{\makecell{$R_{\rm GRPO}$\\RL training}} &
				\multicolumn{2}{c|}{MVTec} &
				\multicolumn{2}{c}{AD-DualDiff} \\
				\cline{5-8}
				& & & 
				& Acc@0.3 & F1-score 
				& Acc@0.3 & F1-score \\
				\hline
				Prompt-only 
				& \xmark & \xmark & \xmark 
				& 38.08 & 30.98 & 40.65 & 23.88 \\
				
				Dual-image Prompt 
				& \cmark & \xmark & \xmark 
				& 43.84 & 32.61 & 41.25 & 26.06 \\
				
				Cold-start Only 
				& \cmark & \cmark & \xmark 
				& 36.32 & 51.28 & 32.20 & 43.16 \\
				
				RL w/o Cold-start 
				& \cmark & \xmark & \cmark 
				& 48.14 & 62.79 & 38.03 & 50.05 \\
				
				RL w/o Cold-start with Single Detection
				& \xmark & \xmark & \cmark 
				& 32.52 & 46.04 & 26.90 & 36.18 \\
				
				DifferAD-R1 with Single Detection
				& \xmark & \cmark$^\dagger$ & \cmark$^\ddagger$ 
				& 53.13 & 55.81 & 52.95 & 49.86 \\
				
				DifferAD-R1 Full 
				& \cmark & \cmark & \cmark 
				& \textbf{70.40} & \textbf{74.68} 
				& \textbf{61.12} & \textbf{60.95} \\
				\hline
			\end{tabular}
		}
		
		\vspace{0.5mm}
		\footnotesize{
			$^\dagger$ The single-image variant is initialized from the same dual-image cold-start checkpoint. 
			$^\ddagger$ RL fine-tuning is performed under the single-image detection setting.
		}
	\end{reviewbluetable}
\end{table}

\begin{table}[h]
	\caption{Ablation study on localization reward design (dual-image paradigm fixed).}
	\label{tab:ablation_reward}
	\centering
	\resizebox{\linewidth}{!}{
		\begin{tabular}{lcccc}
			\hline
			\multirow{2}{*}{Reward Design} &
			\multicolumn{2}{c}{MVTec} &
			\multicolumn{2}{c}{AD-DualDiff} \\
			\cline{2-5}
			& Acc@0.3 (\%) & F1-score (\%) & Acc@0.3 (\%) & F1-score (\%) \\
			\hline
			IoU Reward Only
			& 66.82 & 71.59 & 60.87 & 59.23 \\
			Center-based Reward Only
			& 68.14 & 73.18 & 60.59 & 58.51 \\
			Dual-Consistency Localization Reward (Ours)
			& \textbf{70.40} & \textbf{74.68} & \textbf{61.12} & \textbf{60.95} \\
			\hline
		\end{tabular}
	}
\end{table}

\begin{table}[h]
	\caption{Ablation study on GRPO optimization.}
	\label{tab:ablation_training}
	\centering
	\resizebox{\linewidth}{!}{
		\begin{tabular}{lcccc}
			\hline
			\multirow{2}{*}{Training Strategy} &
			\multicolumn{2}{c}{MVTec} &
			\multicolumn{2}{c}{AD-DualDiff} \\
			\cline{2-5}
			& Acc@0.3 (\%) & F1-score (\%) & Acc@0.3 (\%) & F1-score (\%) \\
			\hline
			GRPO w/o Resampling \& Difficulty
			& 65.79 & 71.57 & 57.47 & 58.59 \\
			+ Group-wise Resampling
			& 67.48 & 73.07 & 60.08 & 59.56 \\
			+ Difficulty-aware Weighting
			& 66.36 & 72.90 & 58.95 & 60.08 \\
			\textbf{Full Strategy (Ours)}
			& \textbf{70.40} & \textbf{74.68} & \textbf{61.12} & \textbf{60.95} \\
			\hline
		\end{tabular}
	}
\end{table}

\begin{table}[h]
	\begin{reviewbluetable}{1-3,3-2,3-3}
	\caption{Hyperparameter sensitivity analysis of DCLR.}
	\label{tab:dclr_hyperparameter_sensitivity}
	\centering
	\resizebox{\linewidth}{!}{
		\begin{tabular}{llcccc}
			\hline
			\multirow{2}{*}{Setting} &
			\multirow{2}{*}{Value} &
			\multicolumn{2}{c}{MVTec} &
			\multicolumn{2}{c}{AD-DualDiff} \\
			\cline{3-6}
			& & Acc@0.3 (\%) & F1-score (\%) & Acc@0.3 (\%) & F1-score (\%) \\
			\hline
			
			Default DCLR
			& $\alpha=0.6,\lambda=0.3,\tau=0.3$
			& \textbf{70.40} & \textbf{74.68} & 61.12 & 60.95 \\
			\hline
			
			\multicolumn{6}{l}{\textit{Sensitivity to $\alpha$ with $\lambda=0.3$ and $\tau=0.3$}} \\
			$\alpha$
			& 0.3 & 67.14 & 73.91 & 58.74 & 59.78 \\
			$\alpha$
			& 0.9 & 65.18 & 72.50 & 56.64 & 59.01 \\
			\hline
			
			\multicolumn{6}{l}{\textit{Sensitivity to $\lambda$ with $\alpha=0.6$ and $\tau=0.3$}} \\
			$\lambda$
			& 0.1 & 68.12 & 74.66 & 60.04 & 61.35 \\
			$\lambda$
			& 0.5 & 65.54 & 72.81 & 58.05 & 59.14 \\
			\hline
			
			\multicolumn{6}{l}{\textit{Sensitivity to $\tau$ with $\alpha=0.6$ and $\lambda=0.3$}} \\
			$\tau$
			& 0.1 & 66.21 & 73.23 & \textbf{60.13} & \textbf{62.01} \\
			$\tau$
			& 0.5 & 65.93 & 73.52 & 59.71 & 60.93 \\
			
			\hline
		\end{tabular}
	}
\end{reviewbluetable}
\end{table}

We conduct comprehensive ablation studies on both the proposed \textbf{AD-DualDiff} dataset and the \textbf{MVTec-AD} benchmark to analyze the effectiveness of key design choices in DifferAD-R1, including the difference paradigm, localization reward, GRPO optimization, RL exploration and sampling scale.

\paragraph{Effect of Difference Paradigm}
\Rev{3-3}{Table~\ref{tab:ablation_paradigm} presents the ablation results of different input and training paradigms.
Using a dual-image prompt without training only brings limited gains over the prompt-only baseline, indicating that simply introducing a reference image is insufficient without task-specific alignment.
The cold-start-only model improves F1-score but still yields limited Acc@0.3, suggesting that natural-image difference pretraining provides useful comparison priors but cannot fully meet the fine-grained localization requirements of industrial anomalies.

When RL is performed without the cold-start stage, performance improves over the cold-start-only setting, but remains clearly below the full model.}
\Rev{1-2,3-3}{Compared with ``RL w/o Cold-start,'' DifferAD-R1 Full improves Acc@0.3/F1-score by 22.26\%/11.89\% on MVTec and 23.09\%/10.90\% on AD-DualDiff, demonstrating that the cold-start phase provides a more stable and effective initialization for subsequent RL optimization.
Finally, compared with the single-image detection baseline, DifferAD-R1 Full achieves substantial gains on both datasets, increasing Acc@0.3/F1-score from 53.13\%/55.81\% to 70.40\%/74.68\% on MVTec and from 52.95\%/49.86\% to 61.12\%/60.95\% on AD-DualDiff.
In addition, the comparison between ``RL w/o Cold-start'' and ``RL w/o Cold-start with Single Detection'' further confirms the advantage of the dual-image paradigm.
Even without cold-start initialization, the dual-image RL variant improves Acc@0.3/F1-score from 32.52\%/46.04\% to 48.14\%/62.79\% on MVTec and from 26.90\%/36.18\% to 38.03\%/50.05\% on AD-DualDiff.
These results validate that explicitly modeling reference-target differences offers more reliable cues for anomaly localization than analyzing the target image in isolation.}


\paragraph{Effect of Localization Reward}
We further investigate different localization reward designs under the paired-image paradigm.
Center-based reward is employed for comparison, which uses the consistency term $S_{\mathrm{cons}}(B_p, B_{gt})$ in DCLR as the reward without jointly incorporating the IoU component.
As showed in Table~\ref{tab:ablation_reward},
Since both the IoU-only and center-based rewards lead to noticeable improvements,
the full DifferAD-R1 model, which integrates dual-consistency localization objectives,
achieves the best performance on both datasets.

\paragraph{Effect of GRPO Optimization}
Table~\ref{tab:ablation_training} analyzes the contribution of different training strategies.
Comparing with the basic GRPO, adding Group-wise Resampling improves performance by increasing the diversity of informative response groups.
Difficulty-aware Weighting further enhances robustness by emphasizing harder localization cases.
When both strategies are jointly applied, DifferAD-R1 achieves the highest F1-score 74.68\% on MVTec-AD and 60.95\% on AD-DualDiff,
highlighting the complementary benefits of resampling and difficulty-aware optimization in reinforcement learning for anomaly localization.

\Rev{1-3,2-2,3-3}{
	\paragraph{Hyperparameter Sensitivity of DCLR}
	Table~\ref{tab:dclr_hyperparameter_sensitivity} reports the sensitivity analysis of the three key hyperparameters in DCLR, including $\alpha$, $\lambda$, and $\tau$. The default setting, $\alpha=0.6$, $\lambda=0.3$, and $\tau=0.3$, is chosen to slightly emphasize center alignment, provide moderate consistency guidance for low-IoU predictions, and align the transition threshold with the Acc@0.3 localization criterion. As shown in the table, this setting achieves the best overall localization accuracy, with 70.40\%/74.68\% Acc@0.3/F1-score on MVTec and 61.12\%/60.95\% on AD-DualDiff. When varying $\alpha$, $\lambda$, or $\tau$, the performance remains within a relatively stable range, although overly emphasizing a single geometric term or using too weak/strong consistency guidance leads to degradation. These results indicate that DCLR is robust to moderate hyperparameter variations and that a balanced combination of IoU, center consistency, and scale consistency provides more stable localization supervision.
}
\begin{figure}[h]
	\centering
	\subfloat[]{%
		\includegraphics[width=0.48\linewidth]{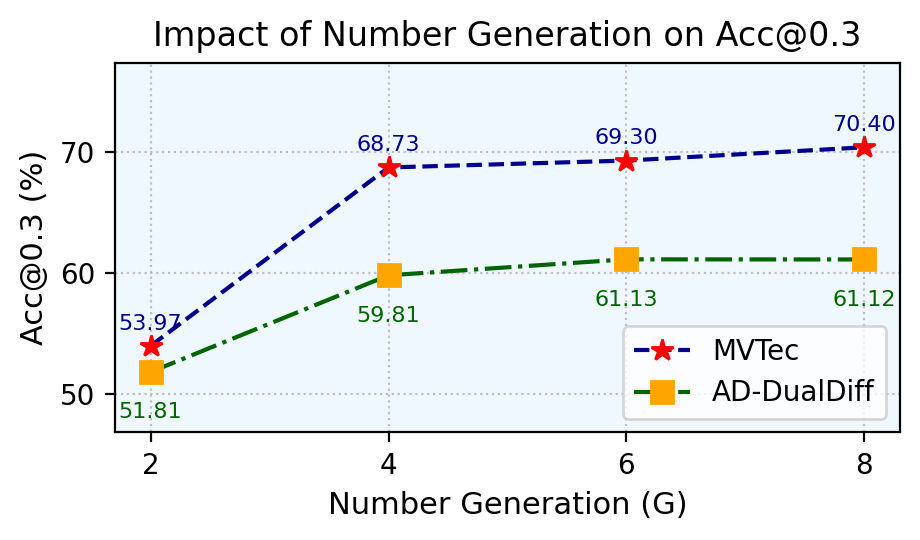}%
		\label{fig:ablation_a}
	}
	\hfil
	\subfloat[]{%
		\includegraphics[width=0.48\linewidth]{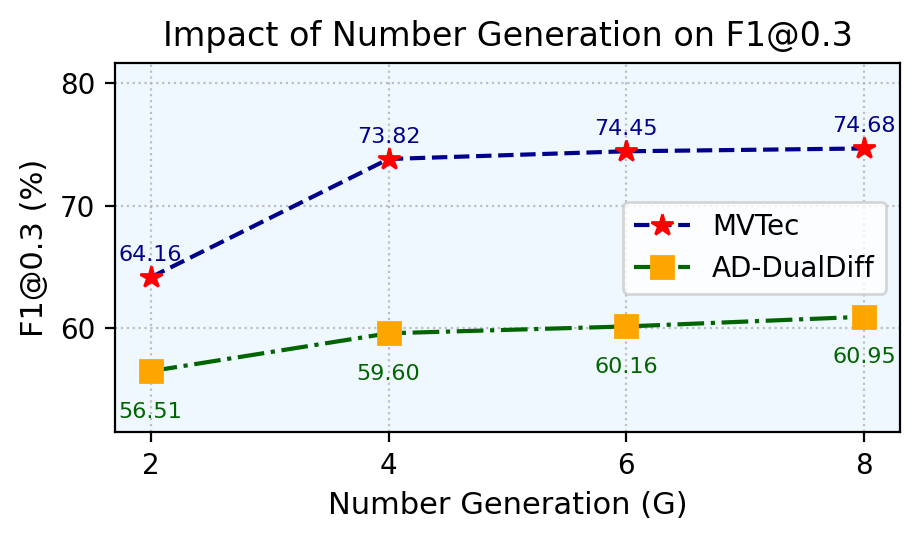}%
		\label{fig:ablation_b}
	}
	
	\vspace{2mm}
	
	\subfloat[]{%
		\includegraphics[width=0.48\linewidth]{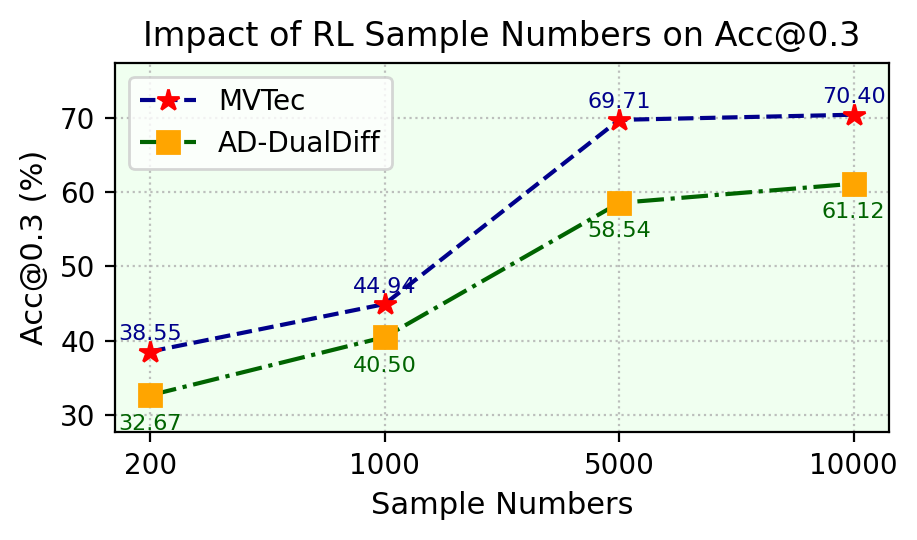}%
		\label{fig:ablation_c}
	}
	\hfil
	\subfloat[]{%
		\includegraphics[width=0.48\linewidth]{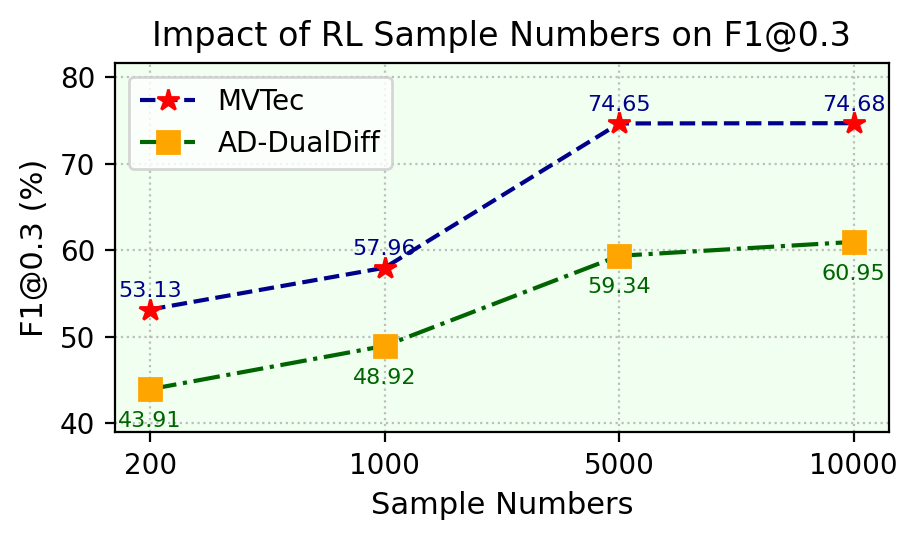}%
		\label{fig:ablation_d}
	}
	
	\caption{Ablation results on reinforcement learning scale.
		(a) Effect of number generation on Acc@0.3.
		(b) Effect of number generation on F1@0.3.
		(c) Effect of RL sample numbers on Acc@0.3.
		(d) Effect of RL sample numbers on F1@0.3.}
	\label{fig:ablation_G_rl_rate}
\end{figure}

\paragraph{Effect of RL Exploration and Sampling Scale}
We analyze the impact of reinforcement learning exploration and sampling scale on anomaly localization by varying the number of generated responses per input ($G$) and the total number of RL training samples (Fig.~\ref{fig:ablation_G_rl_rate}).
Increasing $G$ from 2 to 4 yields significant gains on both MVTec-AD and AD-DualDiff in terms of Acc@0.3 and F1@0.3, as richer response diversity provides more informative relative rewards and stabilizes advantage estimation.
Further increasing $G$ to 6 and 8 brings additional but diminishing improvements, indicating saturation in exploration benefits, we adopt
$G = 8$ as our default setting.

Similarly, increasing the number of RL samples consistently enhances localization performance.
With limited samples, the model fails to adequately capture diverse and hard anomaly patterns, leading to inferior accuracy.
Scaling the sample size to 5,000 and 10,000 results in notable improvements, particularly in F1@0.3, reflecting stronger robustness and recall.
Overall, these results demonstrate that both sufficient exploration and adequate sampling scale are essential for stable and effective reinforcement learning under sparse and noisy localization rewards.

\begin{figure}[t]
	\centering
	\includegraphics[width=3.5in]{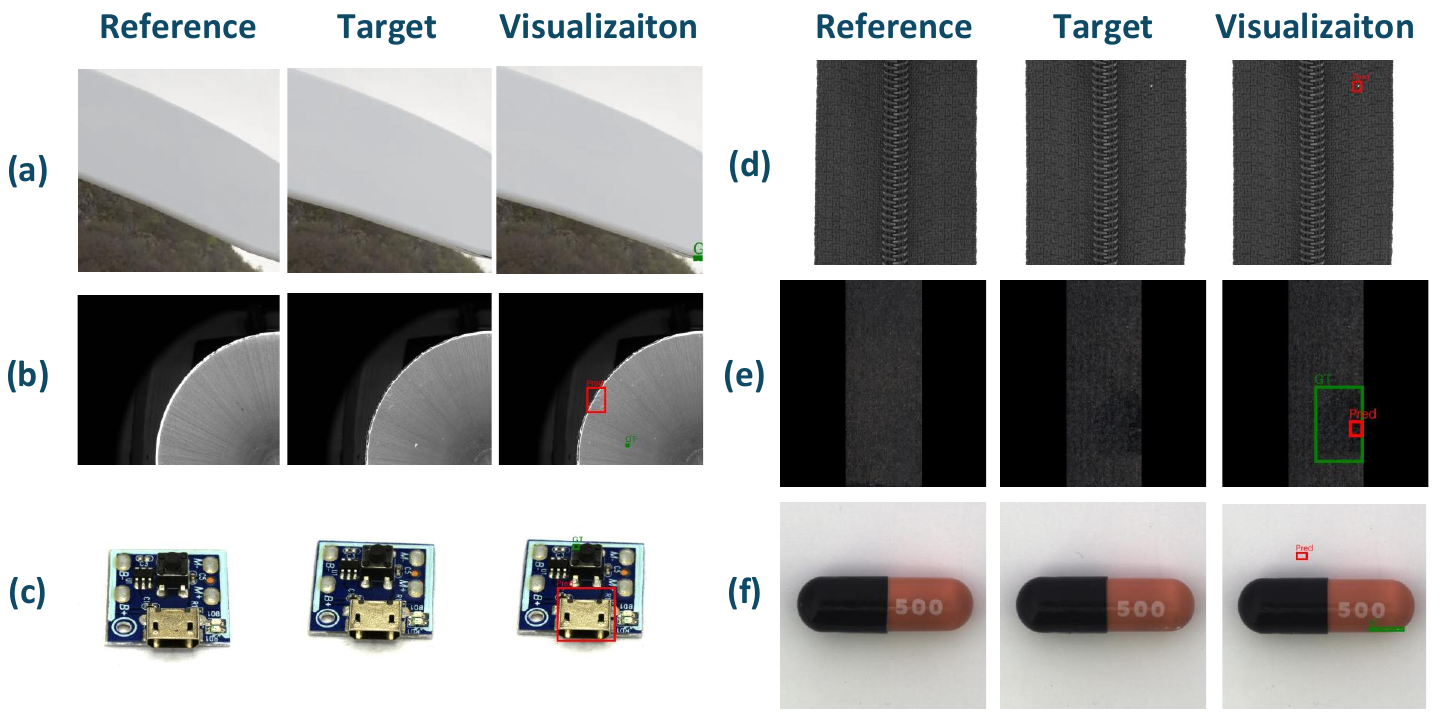} 
	\caption{\Rev{3-5}{Illustration of representative failure cases using DifferAD-R1.}}
	\label{fig:failure_cases}
\end{figure}

\Rev{3-5}{
	\subsection{Discussion}
	
	DifferAD-R1 works effectively because it reformulates industrial anomaly localization as a reference-guided difference grounding task. By comparing a target image with a normal reference image, the model focuses on anomaly-relevant discrepancies rather than category-specific appearances in the target image alone. The two-stage training strategy transfers general cross-image difference perception to industrial anomaly localization, while DCLR and difficulty-aware optimization further stabilize learning for small and subtle defects.
	
	Fig.~\ref{fig:failure_cases} shows representative failure cases. 
	In case (a), slight viewpoint/background variation and a low-contrast defect cause a missed detection. 
	Cases (b), (d), and (f) involve extremely small, thin, or low-contrast defects, where weak visual evidence may lead to false positives or inaccurate localization. 
	Cases (c) and (e) contain highly structured or reflective components, causing the model to focus on salient normal regions or only the most discriminative part of the defect. 
	These results show that DifferAD-R1 may still be affected by subtle or non-defective visual discrepancies, especially when distinguishing true anomalies from weak visual variations requires fine-grained comparison and anomaly-aware reasoning.
}
	
\Rev{3-4}{
	Although DifferAD-R1 is mainly presented as a methodological exploration of difference-grounding for industrial anomaly localization, we have also investigated its practical deployment. Specifically, in a recent industrial project, we migrated the proposed framework to a lightweight Qwen3-VL-2B backbone and deployed it on a robot platform equipped with a single NVIDIA RTX 4090 GPU with 24GB memory. For input image pairs with a resolution of 800$\times$800, the system achieves an inference speed of approximately 100 ms per image pair. In practical industrial inspection, the reference image can often be pre-defined for each product category, while only the target image changes during inspection, which further reduces deployment overhead. These results suggest that the proposed paradigm has practical potential for resource-constrained industrial scenarios. Future work will further improve efficiency through model quantization, knowledge distillation, and parameter-efficient adaptation.
}

\section{Conclusion}

This paper proposes DifferAD-R1, the first work integrating a difference-guided dual-image paradigm with R1-style reinforcement learning for industrial anomaly localization. We design a Dual-Consistency Localization Reward to address GRPO’s limitations and incorporate a difficulty-aware strategy, alongside constructing the AD-DualDiff dataset for real-scenario evaluation. Extensive experiments verify that DifferAD-R1 outperforms existing baselines and achieves competitive performance against large-scale models like Qwen3-VL, highlighting the efficacy of reinforcement learning in hard anomaly localization. Furthermore, we analyze the impact of key modules and hyperparameters, offering valuable insights for future research. By validating the potential of difference-guided reinforcement learning, our work advances the development of industrial anomaly localization toward more robust real-world applications.

For future work, model quantization or knowledge distillation techniques can be explored to achieve economical inference of Multimodal Large Language Models (MLLMs), enabling them to be better adapted to high-speed industrial detection scenarios. In addition, diffusion models can be utilized to generate more realistic defect samples, facilitating in-depth research on industrial anomaly detection methods for multi-defect targets, so as to further improve the model’s adaptability to complex defect scenarios.


\BibRevTag{wang2025Distillation1}{0-1}
\BibRevTag{liu2024unistad3}{0-1} 
\BibRevTag{li2026semaero}{0-1} 
\BibRevTag{xu2025plovad}{0-1} 
\BibRevTag{gu2026filo++}{0-1}
\bibliographystyle{IEEEtran}
\bibliography{reference}

\section*{Biographies}

\begin{IEEEbiography}
	[{\includegraphics[width=1in,height=1.1in,clip,keepaspectratio]{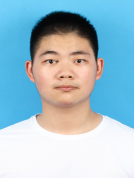}}]
	{Dingrong Wang}
	received the B.S. degree from Northeastern University, China, in 2024.
	He is currently pursuing the M.S. degree with the Institute of Automation,
	Chinese Academy of Sciences, Beijing, China.
	His research interests include computer vision and anomaly detection.
\end{IEEEbiography}

\begin{IEEEbiography}
	[{\includegraphics[width=1in,height=1.1in,clip,keepaspectratio]{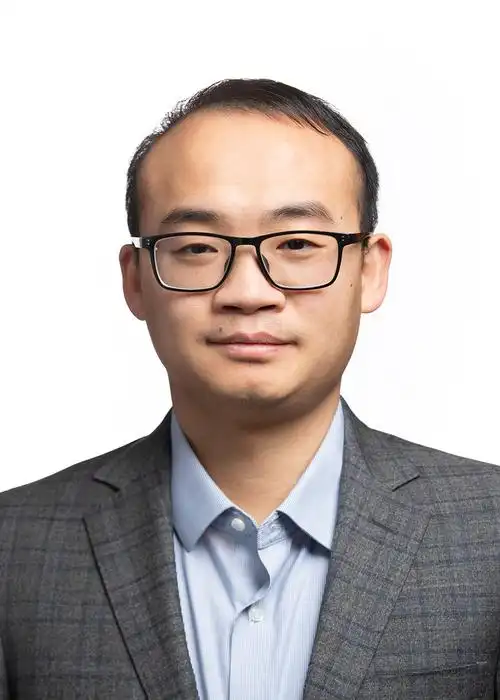}}]
	{Xian Tao}
	(Senior Member, IEEE) received the Ph.D. degree in control theory and
	control engineering from the Institute of Automation,
	Chinese Academy of Sciences, Beijing, China, in 2016.
	He is currently an Associate Professor with the Institute of Automation,
	Chinese Academy of Sciences.
	His current research interests include deep learning and visual inspection.
\end{IEEEbiography}

\begin{IEEEbiography}
	[{\includegraphics[width=1in,height=1.1in,clip,keepaspectratio]{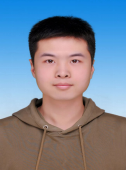}}]
	{Zhen Qu}
	received the B.S. degree from Xidian University, Xi’an, China, in 2022.
	He is currently pursuing the Ph.D. degree with the Institute of Automation,
	Chinese Academy of Sciences, Beijing, China.
	His research interests include machine learning and visual inspection.
\end{IEEEbiography}

\begin{IEEEbiography}
	[{\includegraphics[width=1in,height=1.1in,clip,keepaspectratio]{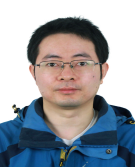}}]
	{Hengliang Luo}
	received the B.Eng. degree in mechanical design, manufacturing,
	and automation and the M.S. degree in agricultural equipment engineering from
	China Agricultural University, Beijing, China, in 2010 and 2013, respectively.
	He is currently working at CASI Vision Technology Co., Ltd., Luoyang, China.
	His research interests include computer vision and machine learning.
\end{IEEEbiography}

\begin{IEEEbiography}
	[{\includegraphics[width=1in,height=1.1in,clip,keepaspectratio]{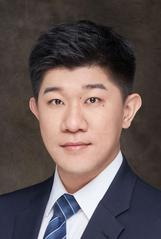}}]
	{Xinyi Gong}
	received the Ph.D. degree in control theory and control engineering
	from the Institute of Automation, Chinese Academy of Sciences, Beijing, China,
	in 2019.
	He is currently a Professor with the Space Information Research Institute,
	Hangzhou Dianzi University, Hangzhou, China.
	His research interests include computer vision, image processing,
	pattern recognition, and machine learning.
\end{IEEEbiography}

\begin{IEEEbiography}
	[{\includegraphics[width=1in,height=1.1in,clip,keepaspectratio]{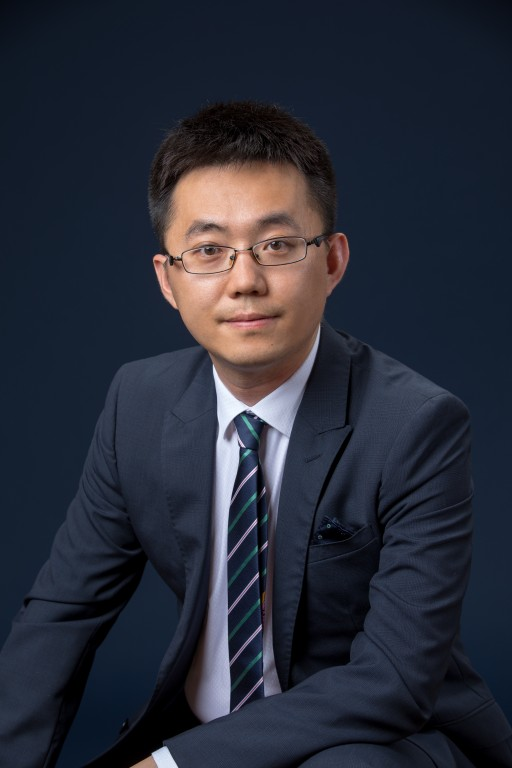}}]
	{Fei Shen}
	received the B.S. degree from Xidian University, Xi’an, China, in 2007,
	the M.S. degree from Beijing Institute of Technology, Beijing, China, in 2009,
	and the Ph.D. degree in control science and engineering from the Institute of
	Automation, Chinese Academy of Sciences, Beijing, China, in 2012.
	He is currently a Professor with the Engineering Laboratory for Intelligent
	Equipment, Chinese Academy of Sciences.
	His research interests include visual inspection, robot vision control,
	and micro-assembly.
\end{IEEEbiography}

\begin{IEEEbiography}
	[{\includegraphics[width=1in,height=1.1in,clip,keepaspectratio]{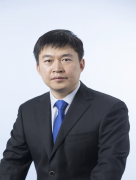}}]
	{Zhengtao Zhang}
	(Member, IEEE) received the Ph.D. degree in control science and
	engineering from the Institute of Automation, Chinese Academy of Sciences,
	Beijing, China, in 2010.
	He is currently a Professor with the Engineering Laboratory for Intelligent
	Equipment and Technology of Industrial Vision, Chinese Academy of Sciences.
	His research interests include visual measurement, micro-assembly,
	and automation.
\end{IEEEbiography}

\begin{IEEEbiography}
	[{\includegraphics[width=1in,height=1.1in,clip,keepaspectratio]{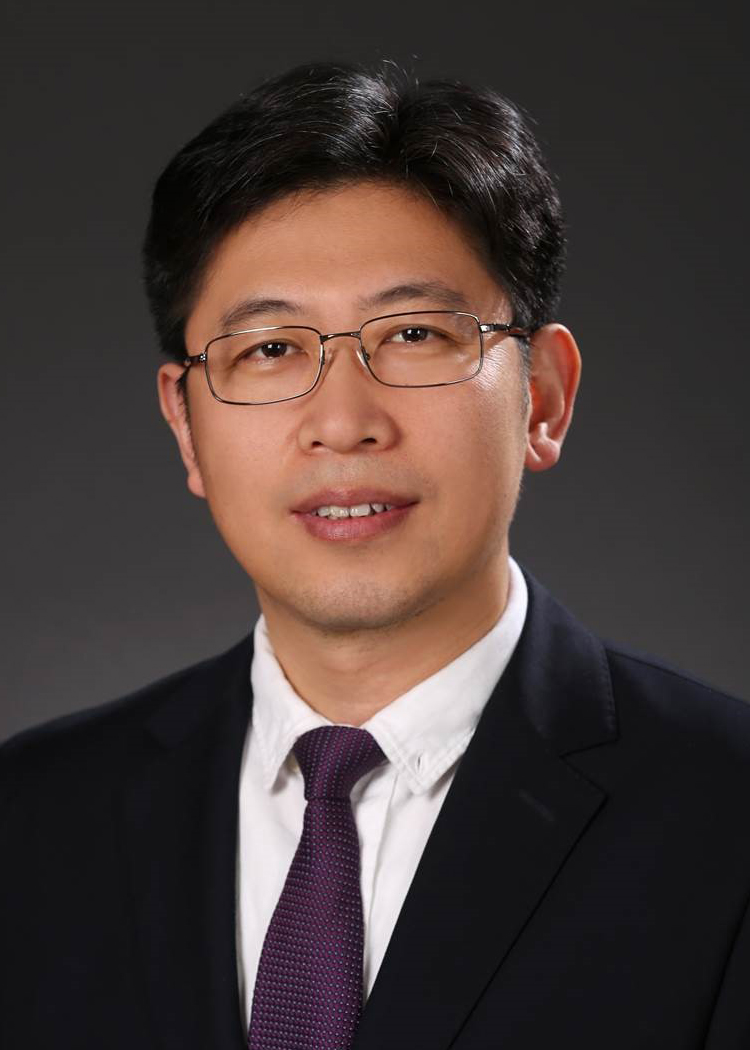}}]
	{Guiguang Ding}
	(Senior Member, IEEE) received the B.E. and Ph.D. degrees from
	Xidian University, Xi’an, China, in 1999 and 2004, respectively.
	He is currently a Professor with the School of Software,
	Tsinghua University, Beijing, China.
	His research interests include multimedia information retrieval,
	computer vision, and machine learning.
\end{IEEEbiography}

\newpage
\vfill

\end{document}